\documentclass{article}
\usepackage{style/iclr2027_conference,times}

\usepackage{amsmath,amsfonts,bm}

\def\eqref#1{equation~\ref{#1}}

\def\1{\bm{1}}

\DeclareMathAlphabet{\mathsfit}{\encodingdefault}{\sfdefault}{m}{sl}
\SetMathAlphabet{\mathsfit}{bold}{\encodingdefault}{\sfdefault}{bx}{n}

\usepackage{hyperref}
\usepackage{url}
\usepackage{booktabs}
\usepackage{multirow}
\usepackage{amsmath}
\usepackage{amsthm}
\usepackage{amssymb}
\usepackage{graphicx}
\usepackage{placeins}
\usepackage{wrapfig}
\usepackage{xcolor}
\usepackage{algorithm}
\usepackage{algpseudocode}

\hypersetup{
  hidelinks,
  hypertexnames=false,
  pdftitle={Self-Generated Feedback Destabilizes Test-Time Training: A Causal Decomposition of Long-Horizon Adaptation},
  pdfauthor={Cheng Luo, Bing Li, Bernard Ghanem}
}

\title{\raggedright
Self-Generated Feedback Destabilizes Test-Time Training:\\[-0.15em]
{\Large A Causal Decomposition of Long-Horizon Adaptation}}
\author{
Cheng Luo \quad
Bing Li\thanks{Corresponding author.} \quad
Bernard Ghanem\\
King Abdullah University of Science and Technology (KAUST)
}
\iclrfinalcopy

\begin{document}
\maketitle

\begin{abstract}
Test-time training (TTT) lets a model store information in its weights during
inference. When the model learns from its own output, however, each update also
changes the model that generates the next training example. Across 128K-token
streams, retaining generated-text updates worsens prediction on independent
human-written text with three TTT-E2E model configurations (labeled 125M,
760M, and 3B). The same
failure occurs when Adam updates Qwen3-4B's existing weights. The same update
mechanisms can improve on real text, so writing itself is not the failure.
Three matched comparisons trace the causal pathway. \emph{Fixed Generation} removes over
$98\%$ of the damage at 125M and 760M by using a frozen model to generate training chunks.
\emph{Recorded Replay} separates the loss caused by reading degraded text from
the additional loss stored by updating on it. A paired one-update comparison
then shows the local conflict: an update predicts its source better but new
real text worse. This cost grows after Closed Loop adaptation, with a few
trajectories accounting for most large failures. Finally, \emph{Settlement}
evaluates the candidate state on independent real text before commitment. It
leaves mean endpoint gaps of $.07$ and $-.02$ nats at 125M and 760M while
retaining real-text adaptation. These results motivate checking prediction on
independent evidence before retaining an update.\footnote{\raggedright
The project repository is publicly available at
{\urlstyle{same}\url{https://github.com/lingjivoo/ttt-ouroboros}}.\par}
\end{abstract}

\section{Introduction}
\label{sec:intro}
Test-time training (TTT) lets part of a model continue to learn during
inference \citep{sun2020ttt,tttplus,tttlayers}. In long-context language
models, these changing parameters---the \emph{fast weights}---can retain
information after the source text leaves attention
\citep{schmidhuber1992fast,ba2016fast,schlag2021linear,tttE2E,titans}. This
makes TTT attractive for long documents, streams, and long-running agents,
where retaining the full history in attention is costly
\citep{tttE2E,titans,agenticttt}. We call processing text without changing
weights \emph{reading}, and retaining its update \emph{writing}.

Writing becomes risky when the model also supplies its future training data.
Fast weights $W_t$ generate chunk $x_t$, learning from it produces $W_{t+1}$,
and the new weights generate the next chunk. Lower loss on $x_t$ shows that the
update fits its source, not that it predicts other text better. We ask which
part of this feedback loop makes persistent self-writing fail.

Over 128K-token streams, matched models retain or discard each generated-text
update and are evaluated on separate human-written passages. In End-to-End
Test-Time Training (TTT-E2E) \citep{tttE2E}, retaining generated-text writes
worsens real-text prediction at 125M, 760M, and 3B. The effect emerges beyond
the original evaluation horizon and also occurs when Adam updates Qwen3-4B's
existing feed-forward weights \citep{kingma2015adam,qwen3}; it is not specific
to TTT-E2E's update module. The same update rules improve prediction when
trained on real text, so the question is why closed-loop generation reverses
their effect.

We trace the causal pathway through three matched
comparisons. \textbf{First, Fixed Generation removes feedback to future
training text:} a frozen model generates each chunk while a separate model
learns from it, removing over $98\%$ of the damage at 125M and 760M. \textbf{Second, Recorded
Replay holds the text fixed:} degraded tokens hurt through attention, and
retaining their updates adds a persistent weight cost, including at 3B.
\textbf{Third, paired one-update branches start from identical weights,
attention, and text:} retaining the update improves its source prediction but
worsens the next real passage. This cost grows after Closed Loop adaptation,
and a few trajectories produce most large failures.

A repetition penalty slows the feedback trajectory, reducing the 128K endpoint
gap from $3.01$ to $.88$ nats, but repetition does not identify which passage
is safe to write. \emph{Settlement} instead evaluates the candidate state on
independent real text. It preserves real-text adaptation and leaves mean
endpoint gaps of $.07$ and $-.02$ nats at 125M and 760M, respectively.
It also remains stable when source labels are noisy.

Recursive synthetic-data studies track feedback across successive model
generations \citep{selfconsuming,shumailov2024collapse}. aTTT studies
feedback within an agent episode and downweights repeated update tokens
\citep{agenticttt}; VANE validates updates on later visual observations
\citep{vane}. Here we isolate feedback in long language-model streams, using
matched comparisons to distinguish changes to future training text from the
effects of reading and updating on fixed text. Our contributions are:
\textbf{(1) long-horizon evidence} that
self-generated feedback damages independent real-text prediction across three
TTT-E2E scales and under ordinary Adam, together with an external-text exposure
boundary; \textbf{(2) a causal decomposition} using Fixed Generation, Recorded
Replay, and paired one-update branches to separate changes to future training
text, attention, and persistent weights, revealing a state-dependent failure
concentrated in a few trajectories; and \textbf{(3) a transfer criterion} showing why
source fit and repetition cannot justify retaining an update, and how
independent text can evaluate the complete candidate state.

\section{Preliminaries: Persistent Test-Time Training}
\label{sec:preliminaries}

\paragraph{TTT-E2E learns during inference.}
An ordinary language model keeps the same parameters throughout inference.
Test-time training (TTT) instead updates part of the model on the input it is
currently processing, allowing information to remain in the weights after the
corresponding tokens leave the attention window. Applying such updates only at
inference creates a mismatch with pretraining. TTT-E2E addresses this mismatch
by applying the same next-token updates during training and meta-learning an
initialization that works well after those updates \citep{tttE2E}. At inference,
it updates MLP blocks in the final quarter of a sliding-window Transformer
\citep{vaswani2017attention,tttE2E}. Following the fast-weight literature
\citep{schmidhuber1992fast,ba2016fast}, we denote these adaptable parameters by
$W_t$: $W_0$ is their initial value and $W_t$ is their value after $t$ chunks.

\paragraph{Reading, writing, and the feedback loop.}
Every chunk can affect later predictions by remaining in attention; we call
this \emph{reading}. If the model also retains the gradient update learned from
that chunk, the chunk changes $W_t$; we call this \emph{writing}. When the model
learns from its own output, the current state $W_t$ generates chunk $x_t$, the
update on $x_t$ produces $W_{t+1}$, and that new state helps generate the next
chunk. This recurring dependence is the \emph{Closed Loop}. Our main control,
\emph{Generated Writes Off} (hereafter \emph{Writes Off}), reads the same kind
of generated chunks but discards the update after each one. Both policies keep
updates from the initial real-text prefix. Their difference is therefore
whether learning from generated text becomes persistent, not whether the model
sees generated text at all.

\section{Experimental Setup}
\label{sec:setup}
\paragraph{Models and updates.}
TTT-E2E pretrains on DCLM and continues long-context training on books
\citep{dclm,tttE2E}. We follow that domain setting with PG-19
\citep{pg19}: we train models labeled 125M and 760M by their TTT-E2E
configurations and use
the released 3B model. Most causal experiments use the 125M model. It has
an 8192-token attention window, and one clipped gradient step follows each
1024-token chunk, updating feed-forward modules in the final three layers. To
check that the phenomenon is not specific to TTT-E2E, we also apply Adam
\citep{kingma2015adam} directly to existing Qwen3-4B weights \citep{qwen3}.
The main cross-scale comparison uses the same six books, five seeds, 128K
horizon, and batch width eight at every scale.
Appendix~\ref{app:protocol-map} gives the complete configuration map.

\paragraph{Streams and independent evaluation.}
Each 128K stream starts with 8K tokens of real text. The model then generates
105 chunks, with 15 later real-text evaluations plus an initial reference. In
the canonical comparison, these passages come from later, disjoint regions of
the same PG-19 book. After each evaluation, we restore the preceding weights
and attention state, so evaluation cannot change the continuing stream.
Throughout the paper, \emph{independent real text} means text outside the
generated training stream; it need not come from a different book. Its NLL
asks whether an update transfers beyond the text that produced it. It does not
measure whether the model remembers its own earlier output. K denotes 1024
tokens. Generation uses temperature one and top-$p=0.95$ nucleus sampling
\citep{holtzman2020curious}.
Appendix~\ref{app:implementation} gives complete schedules and
state-restoration checks.

\paragraph{Primary comparison and outcomes.}
We want to measure deterioration caused by retaining generated-text updates,
not the change that occurs even when those updates are disabled. Both policies
therefore start from identical weights and an identical real-text prefix. The
Closed Loop retains each generated-text update; Writes Off discards it, while
both policies continue to read generated tokens. For condition $c$, let
$L_{c,\mathrm{first}}$ and $L_{c,\mathrm{last}}$ be NLL on its first and last
independent real-text evaluations. Lower NLL is better. We define
\begin{equation}
D_c=L_{c,\mathrm{last}}-L_{c,\mathrm{first}},
\qquad H_c=D_c-D_{\mathrm{off}}.
\label{eq:damage}
\end{equation}
$D_c$ is the first-to-last change within condition $c$: positive values mean
that real-text prediction became worse. $D_{\mathrm{off}}$ is the same change
under Writes Off. Their difference, $H_c$, is the additional deterioration
caused by the full policy relative to disabling generated-text writes. Thus
$H_c>0$ means worse than Writes Off, and $H_c<0$ means better.

When both policies receive identical prerecorded tokens, $H_c$ isolates the
effect of retaining their updates.
During online generation, it includes both the updates and the later text that
those updates cause the model to generate; this total feedback effect is our
primary outcome. Intervention suites additionally report the endpoint gap,
$E_c=L_{c,\mathrm{last}}-L_{\mathrm{off},\mathrm{last}}$. Positive $E_c$
means that damage remains at the end; zero means that the final NLL matches
Writes Off. When initial losses match, $E_c=H_c$. Unless specified otherwise,
95\% intervals resample books after averaging seeds within each book.
Appendix~\ref{app:protocol-map} maps every comparison to its configuration and
baseline.

\section{Long-Horizon Persistent Self-Writing Degrades Prediction}
\label{sec:failure}
\begin{wraptable}{r}{.51\linewidth}
\caption{\textbf{Generated-text writes worsen real-text prediction.}
The same six books, five seeds, and 128K horizon are used at every scale.}
\label{tab:current-results}
\centering\scriptsize\setlength{\tabcolsep}{3.2pt}
\begin{tabular}{lrr}
\toprule
Model & $H$ ($\uparrow$ worse) & 95\% Interval\\
\midrule
125M TTT-E2E & $+3.01$ & $[+1.86,+4.20]$\\
760M TTT-E2E & $+6.00$ & $[+4.27,+7.73]$\\
3B TTT-E2E & $+0.40$ & $[+0.30,+0.51]$\\
\bottomrule
\end{tabular}

\end{wraptable}
We first ask whether persistent self-writing eventually makes the model worse
at predicting human-written text. Closed Loop and Writes Off both generate for
128K tokens and keep the generated tokens in attention. Closed Loop alone
retains the update after each chunk. We periodically pause both streams and
evaluate the current weights on the same independent real passages. The
difference therefore measures the consequence of making generated-text
updates persistent, beyond merely reading the generated text.

Table~\ref{tab:current-results} uses the same six books at every scale.
Retaining the updates adds
$3.01$, $6.00$, and $.40$ nats over Writes Off at 125M, 760M, and 3B,
respectively. Every interval excludes zero, establishing the same directional
failure at all three evaluated scales. The magnitude differs across
models; the result is a cross-scale replication of direction rather than
a monotone relation with parameter count.

\begin{figure}[!t]
\centering
\includegraphics[width=\linewidth]{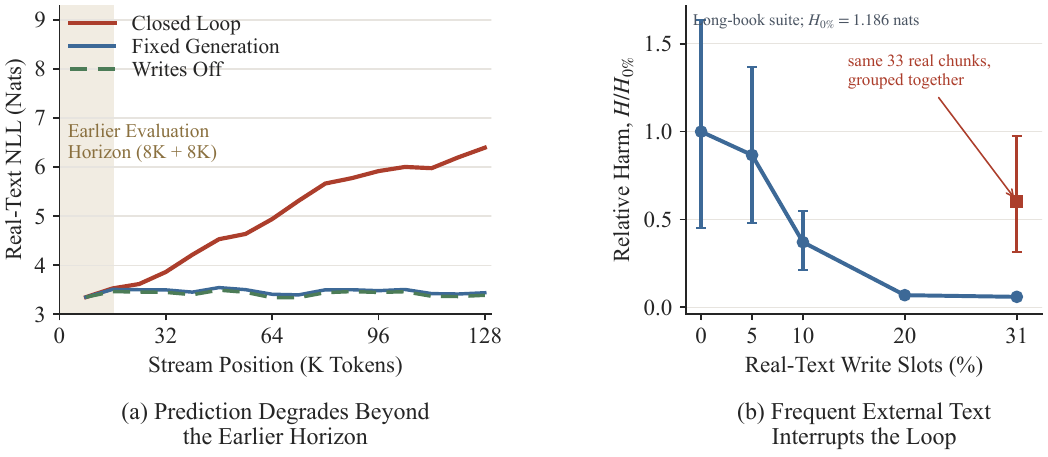}
\caption{\textbf{Long horizons and sparse real-text exposure expose the failure.}
(a) Real-text NLL across a 128K stream; shading marks the earlier 8K+8K
horizon \citep{tttE2E}. (b) Harm relative to a separate long-book suite's 0\%
baseline. Even spacing interrupts the loop; grouping the same 31\% real-text
budget into bursts restores much of the harm. Panel (b) is normalized within a
separate matched suite; error bars scale its 95\% intervals by the fixed
0\%-exposure estimate (Appendix~\ref{app:exposure-density}).}
\label{fig:failure-story}
\end{figure}

Figure~\ref{fig:failure-story}a shows why the long horizon matters. Within the
earlier 8K+8K evaluation window, Closed Loop and Writes Off remain close. Only
after many additional self-writes does Closed Loop rise while Writes Off stays
nearly flat. The failure is therefore cumulative rather than a large penalty
from the first update.

Could decoding alone explain the result? It does not. Full-support sampling
reduces the 128K endpoint gap from $3.010$ to $1.452$ $[.708,2.180]$, but late loss is
still rising. Lower temperature, typical sampling, and a repetition penalty
change the speed or severity of degradation without providing a reliable
boundary between safe and harmful writes. Periodically resetting the fast
weights is stronger: an eight-chunk reset leaves a $.175$ gap $[.084,.321]$, but
also preserves only 44.1\% of the benefit obtained from real-text adaptation.
The full decoder and reset comparisons appear in
Appendix~\ref{app:decoder-controls}.

\paragraph{Frequent external text interrupts the feedback loop.}
Can occasional real text stop the damage from accumulating? We replace 0, 5,
11, 21, or 33 of the stream's 105 generated chunks with real chunks while
retaining every update. Figure~\ref{fig:failure-story}b reports the remaining
harm relative to no replacements. Evenly spaced real chunks reduce it to
approximately 87\%, 37\%, 7\%, and 6\%, respectively.

Timing also matters. With 33 evenly spaced real chunks, the model generates at
most three chunks consecutively. Grouping the same 33 chunks into bursts
permits stretches of 12 and leaves about 60\% of the original harm rather than
6\%. Real text works best when it repeatedly interrupts self-writing
(Appendix~\ref{app:exposure-density}).

\FloatBarrier

\begin{table}[!ht]
\caption{\textbf{Qwen3-4B with Adam.} The same weights benefit from real text
but fail under persistent self-writing. Sampling units and complete settings:
Appendix~\ref{app:inplace}.}
\label{tab:qwen-main}
\centering\small\setlength{\tabcolsep}{9pt}
\begin{tabular}{lrr}
\toprule
Adam LR & \shortstack{Persistent self-writing $H$\\95\% CI; $\uparrow$ worse} &
\shortstack{Real-text $B$\\state range; $\uparrow$ better}\\
\midrule
$10^{-5}$ & $.026\;[-.071,.122]$ & $.139\;[.115,.163]$\\
$10^{-4}$ & $\mathbf{1.231\;[.580,2.350]}$ & $.168\;[.132,.204]$\\
\bottomrule
\end{tabular}
\end{table}
\paragraph{Beyond native TTT: Qwen3-4B shows the same failure.}
We use Adam \citep{kingma2015adam} to update Qwen3-4B's existing feed-forward
weights \citep{qwen3}, without adding a TTT module. At learning rate
$10^{-4}$, persistent self-writing raises real-text NLL by $1.231$ nats relative to
Writes Off; at $10^{-5}$, the interval includes zero
(Table~\ref{tab:qwen-main}). Yet the same $10^{-4}$ update lowers NLL by
$.168$ nats when it learns from real text. Ordinary Adam updates can therefore
help on real text and fail in a self-generated loop: the failure is not
specific to TTT-E2E's learned update mechanism.

\paragraph{Real-text writes can be useful.}
These results do not imply that all inference-time learning should be disabled.
On eleven identical teacher-forced real-text streams, retaining updates lowers
mean NLL by $.0340$ nats relative to \emph{No Updates}
(Appendix~\ref{app:payoff-final}). Because the input text is fixed, the gain
comes from learning rather than from changing future inputs. Writing can help;
the next section asks why it becomes damaging when the learner also supplies
the text used for later updates.

\FloatBarrier

\section{A Causal Decomposition of the Feedback Path}
\label{sec:causal}
\begin{figure}[!ht]
\centering
\includegraphics[width=.97\linewidth]{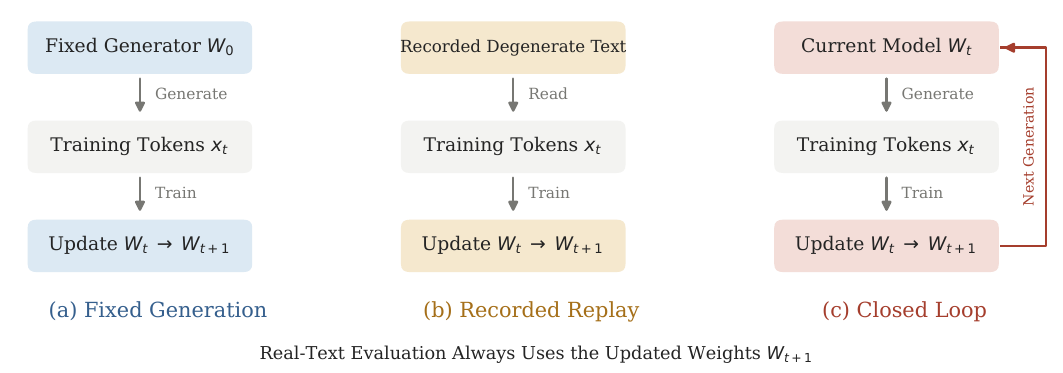}
\caption{\textbf{Three controlled comparisons separate generation, content, and updating.}
(a) The initial state $W_0$ generates while a separate $W_t$ learns, breaking feedback.
(b) Saved tokens are processed read-only or with updates, holding content
fixed. (c) Current $W_t$ both generates and learns, closing the loop. Every
write-enabled condition evaluates the adapted $W_t$; Appendix~\ref{app:control-audit}
verifies generator isolation.}
\label{fig:feedback-paradigms}
\end{figure}

Why does Closed Loop fail? There are three distinct possibilities. Generated
text may be poor training data even when it comes from a fixed model; degraded
tokens may hurt while they remain in attention; and learning from those tokens
may store additional harm in the weights. Figure~\ref{fig:feedback-paradigms}
separates these possibilities by changing one dependency at a time. In
\textbf{Fixed Generation} (a), frozen $W_0$ generates every chunk for a
separate learner, so learning cannot change later training text. In
\textbf{Recorded Replay} (b), a new receiver processes the same saved degraded
tokens with and without updates. In the \textbf{Closed Loop} (c), current
$W_t$ generates $x_t$, learns from it, and becomes $W_{t+1}$ before generating
again. We call this sequence of controlled comparisons a \emph{causal
decomposition}: it locates the path to damage, rather than assigning additive
shares of one total loss. Every comparison evaluates the adapting model on
independent real text.
\subsection{Keep learning, but fix the generator}
\label{sec:separate}
We first ask whether generated text is harmful merely because it is synthetic.
In panel (a), the learner still updates after every chunk, but an independently
cached $W_0$ supplies all future chunks. We evaluate the learner, not $W_0$.

Fixed Generation leaves only $.052$ and $.073$ nats of excess change at 125M
and 760M, removing $98.3\%$ and $98.8\%$ of their matched Closed Loop gaps
(Table~\ref{tab:fixed-scale-main}). The learner still sees generated text and
still updates after every chunk. The only removed link is its ability to
change the text from which it will learn next. Generated origin alone is
therefore insufficient; feedback to future training text accounts for nearly
all of the measured gap in these comparisons.

\begin{table}[!ht]
\caption{\textbf{Breaking feedback removes most of the excess change at both
evaluated scales.} Fixed Generation residuals and removal fractions use matched
within-scale comparisons on the same six canonical books. Complete endpoints
are in Appendix~\ref{app:new-scale}.}
\label{tab:fixed-scale-main}
\centering\small\setlength{\tabcolsep}{5pt}
\begin{tabular}{lrr}
\toprule
Model & Fixed Generation $H$ & Closed Loop harm removed\\
\midrule
125M & $.052$ & $98.3\%$\\
760M & $.073$ & $98.8\%$\\
\bottomrule
\end{tabular}

\end{table}

Update magnitude does not explain the result. Fixed Generation moves farther
from the initial weights than Closed Loop ($.01387$ versus $.00857$) but causes
far less damage. Real-Text Learning moves farther still ($.01759$) and improves
NLL by $.3544$ nats $[.1236,.6048]$. Generated origin is also insufficient:
Fixed Generation learns from generated text without closing the feedback loop.
The distinguishing feature is therefore that the updated weights determine the
text used for later updates.

Fixed Generation necessarily changes the generated sequence, so it cannot tell
us what happens after a degraded sequence has already been produced. Closed
Loop ends almost entirely repetitive (repeated-4 $=.9902$), unlike Writes Off
($.0217$) and Fixed Generation ($.0082$), but that correlation does not separate
the effect of reading the degraded text from the effect of learning from it.
Recorded Replay holds the tokens fixed to separate the two.

\subsection{Recorded Replay: process the same text with and without writing}
A receiver adapted on different books processes a saved degraded stream that
it did not generate. From the same starting state, it first reads the recording
without updating. We then restore the state and process the identical tokens
again while retaining their updates. The first comparison measures the cost of
keeping degraded tokens in attention; the difference between the two passes
measures the additional cost stored in the weights.

\begin{table}[!ht]
\caption{\textbf{Degraded content harms through both attention and weights.}
The 125M read-only excess shows that the recording is harmful in attention;
the paired read-and-write difference measures the additional persistent cost.}
\label{tab:replay-main}
\centering\small\setlength{\tabcolsep}{7pt}
\begin{tabular}{llr}
\toprule
Scale & Receiver operation on the same recorded text & Real-text NLL above control\\
\midrule
125M & Read only & $1.606$\\
125M & Read and write & $3.874$\\
125M & \textbf{Additional cost of writing} & $\mathbf{2.268}$\\
\midrule
3B & \textbf{Additional cost of writing} & $\mathbf{.193\;[.104,.327]}$\\
\bottomrule
\end{tabular}

\end{table}

At 125M, reading alone adds $1.606$ nats over the receiver's control, so the
degraded content is already harmful while it remains in attention. Reading and
writing adds $3.874$ nats; the extra $2.268$ nats is the cost stored by the
weight updates. A 3B replication measures the same write-minus-read
contrast across six source and six disjoint
receiver books. Writing adds $.1927$ $[.1043,.3272]$ nats, and every source-
and receiver-book aggregate is positive
(Appendix~\ref{app:replay-3b-confirmation}).

\begin{figure}[!ht]
\centering
\includegraphics[width=\linewidth]{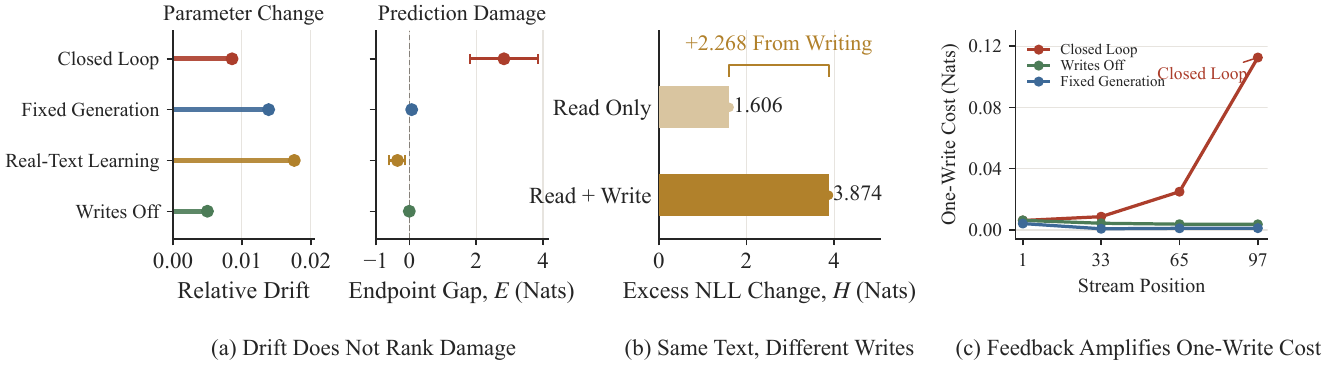}
\caption{\textbf{Feedback, not parameter drift, amplifies the cost of writing.}
(a) Matched 125M control: the same four policies are ordered across relative
drift and final clean-text NLL; horizontal bars are 95\% book intervals.
(b) Identical recorded tokens with updates disabled or retained; only this
panel holds text fixed. Both panels use eight books and five seeds.
(c) In a separate paired one-update comparison, the real-text cost of retaining
one update rises sharply over a Closed Loop history, but stays near zero when
earlier generated writes were discarded or came from Fixed Generation.}
\label{fig:controls-story}
\end{figure}

Figure~\ref{fig:controls-story} links the three results. Panel (a) shows that
parameter movement does not rank damage. Panel (b) holds degraded text fixed
and isolates the added harm from writing it. Panel (c) follows one-update cost
through the stream: it rises sharply only after Closed Loop adaptation and
stays near zero under Writes Off and Fixed Generation. Feedback therefore
changes both future training text and the cost of later writes.

\subsection{One update can fit its source and hurt another input}
\label{sec:single}
The preceding endpoints combine many writes, so they do not show what one
update itself changes. We isolate that effect by copying the same weights and
attention cache immediately before a generated chunk. Both copies process the
same chunk; one keeps its update and the other discards it. Any later difference
is therefore caused by retaining that single update. We measure whether it
improves prediction of its source chunk, whether that improvement transfers to
the next independent real passage, and whether later generation becomes more
repetitive. Table~\ref{tab:single-main} reports \emph{keep minus skip} for the
three outcomes.

The three rows describe the state before this paired update. \emph{Writes Off
History} means all earlier generated-text writes were discarded;
\emph{Closed Loop} means they were retained; and \emph{Fixed Generation}
means earlier updates were learned from chunks generated by frozen $W_0$.
Negative training-chunk NLL means that keeping the update fits its source
better; positive next-real-text NLL means that the same update harms the new
passage.

\begin{table}[!ht]
\caption{\textbf{An update can improve source fit while harming new text.}
125M; keep minus discard, with 95\% book intervals.}
\label{tab:single-main}
\centering\small\setlength{\tabcolsep}{3pt}
\begin{tabular}{lrrr}
\toprule
Preceding History & \shortstack{Training-Chunk NLL\\($\downarrow$ better fit)} & \shortstack{Next Real-Text NLL\\($\uparrow$ worse)} & \shortstack{Future Repeated-4\\($\uparrow$ more repetition)}\\
\midrule
Writes Off History & $-.1374\;[-.1453,-.1285]$ & $+.0045\;[.0037,.0053]$ & $+.0049\;[-.0079,.0248]$\\
Closed Loop & $-.0354\;[-.0384,-.0317]$ & $+.0381\;[.0098,.0718]$ & $+.0086\;[.0020,.0184]$\\
Fixed Generation & $-.1449\;[-.1482,-.1407]$ & $+.0018\;[.0014,.0023]$ & $+.0017\;[-.0034,.0063]$\\
\bottomrule
\end{tabular}

\end{table}

In every row, retaining the update lowers mean source-chunk NLL but raises
mean NLL on the next real passage. Better source fit therefore does not
guarantee a benefit on new text. Fixed Generation has the largest source-fit
improvement ($0.1449$) and the smallest real-text cost ($0.0018$), but even
that cost is positive. Closed Loop has a smaller source improvement
($0.0354$) and a much larger real-text cost ($0.0381$). Within Closed Loop, the
mean cost also grows from $0.006$ nats at the earliest measured position to
$0.113$ at the latest. Both the model state and its generated text change over
this history; the comparison measures their combined effect on later writes.

\begin{table}[!ht]
\caption{\textbf{Two gradient signals rank harmful updates.}
Entries are Spearman correlations with the measured increase in next-real-text
NLL after keeping rather than discarding one update. (a) Negative $\rho$ means
that more-negative gradient cosine predicts greater harm. (b) Positive $\rho$
means that a larger first-order predicted loss increase matches greater measured
harm. Each model-history setting contains 96 paired updates.}
\label{tab:gradient-conflict-main}
\begin{minipage}[t]{.48\linewidth}
\centering
\textbf{(a) Cosine vs. measured harm}\par\smallskip
\scriptsize\setlength{\tabcolsep}{4pt}
\resizebox{\linewidth}{!}{%
\begin{tabular}{lrr}
\toprule
History & 125M $\rho$ & 760M $\rho$\\
\midrule
Closed Loop & $-.502\;[-.642,-.305]$ & $-.633\;[-.808,-.383]$\\
Writes Off & $-.788\;[-.826,-.720]$ & $-.731\;[-.810,-.666]$\\
\bottomrule
\end{tabular}
}
\end{minipage}\hfill
\begin{minipage}[t]{.48\linewidth}
\centering
\textbf{(b) Predicted vs. measured harm}\par\smallskip
\scriptsize\setlength{\tabcolsep}{7pt}
\begin{tabular*}{\linewidth}{@{\extracolsep{\fill}}lrr}
\toprule
History & 125M $\rho$ & 760M $\rho$\\
\midrule
Closed Loop & $+.704$ & $+.632$\\
Writes Off & $+.794$ & $+.751$\\
\bottomrule
\end{tabular*}
\end{minipage}

\end{table}
\paragraph{Gradient conflict predicts which updates are more harmful.}
Table~\ref{tab:gradient-conflict-main} asks a simple question: before keeping an
update, can we tell whether learning the generated chunk will conflict with
predicting new real text? We score each candidate before applying it, then use
the controlled comparison from Table~\ref{tab:single-main}: two identical copies
keep or discard the update and predict the same real passage. The Keep copy's
extra loss is the measured harm.

For panel (a), we compute the cosine between the generated-chunk gradient and
the real-passage gradient for each candidate, then correlate the 96 cosine
scores with their 96 measured harms. The correlations are $-.502/-.788$ after
Closed Loop/Writes Off history at 125M and $-.633/-.731$ at 760M; all four
intervals exclude zero. Thus greater gradient opposition accompanies greater
harm. For panel (b), we compute $g_{\rm real}^{\top}\Delta W$ for each candidate
and correlate these predicted loss changes with the same measured harms. The
correlations are $+.704/+.794$ at 125M and $+.632/+.751$ at 760M. These are
rank correlations, not NLL increases. Both calculations consistently identify
the same pattern: candidates with stronger generated--real conflict cause more
real-text harm. This explains poor transfer at one update; it is not a selection
rule in these experiments: it uses the real passage on which harm is measured
(Appendix~\ref{app:gradient-conflict}).

\paragraph{What the causal decomposition establishes.}
The comparisons establish: \textbf{(1) Feedback:} Fixed Generation prevents
updates from changing later text and removes most damage. \textbf{(2) Reading
and writing:} Replay holds tokens fixed and separates harm from reading them
from the additional harm stored in weights. \textbf{(3) Transfer:} one update
can fit its source yet harm new real text, and gradient conflict ranks that
harm. The last comparison also shows that average one-update cost grows after
Closed Loop adaptation. We next ask whether most late writes become harmful or
whether a small subset drives that increase.

\FloatBarrier

\section{Which Updates Should Be Retained?}
\label{sec:control}
\subsection{Damage amplifies in degraded states but remains heterogeneous}
\label{sec:repairs}
The average one-update cost rises later in Closed Loop. Does this increase
affect most passages, or come from a small harmful subset?
\begin{figure}[H]
\centering
\includegraphics[width=.72\linewidth]{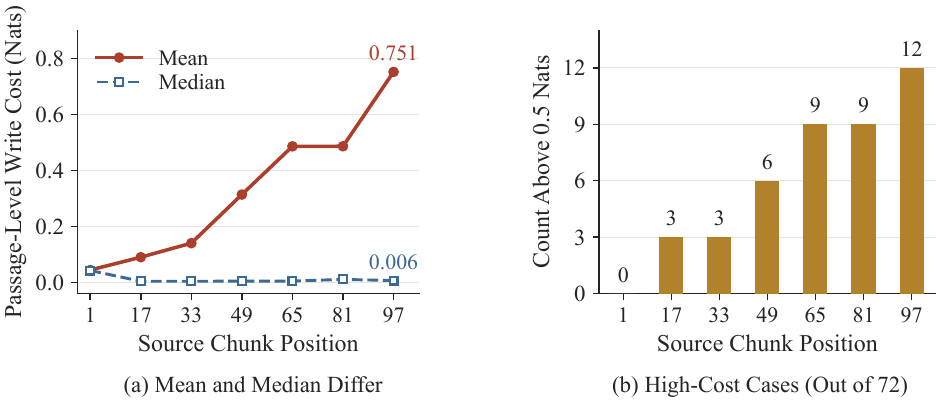}
\caption{\textbf{Damage concentrates in a few late passages.}
(a) Mean and median passage-level write cost; (b) number of costs above
$0.5$ nats among 72 paired comparisons at each position.}
\label{fig:stage-story}
\end{figure}

For each passage, identical receivers process the same tokens as Read Only or
Read+Write, then predict the same real text. We define
\emph{write cost}=NLL(Read+Write)$-$NLL(Read Only); a positive value is harm
caused by retaining the updates, beyond reading the same tokens.
Each position has 72 paired costs
(Appendix~\ref{app:stage-sweep}).

At the first position, mean and median costs are similar ($0.0439$ and
$0.0422$), with no value above $0.5$ nats. At the final position, the mean rises
to $0.7512$ but the median falls to $0.0064$: the typical cost does not increase
with the mean. Only 12 of 72 late costs exceed $0.5$. All twelve come from four
of 24 source sequences. After a source--receiver pair first exceeds the
threshold, its later sampled passages remain above it. Thus a few source
sequences repeatedly produce high-cost passages and raise the late mean; position alone
cannot identify a safe write.

Within repetition bins, the estimated early-to-late cost change is $+.0587$
$[-.0382,.3286]$; this comparison does not resolve a change. The intervention
results provide more direct evidence: repetition weighting and
read-only real context leave $54\%$ and $48\%$ of the Closed Loop gap, whereas
an eight-chunk reset removes $94.2\%$ of the canonical gap but retains only $44.1\%$ of
useful adaptation. These results expose a trade-off: milder controls leave
substantial damage, whereas the stronger reset also removes most of the
adaptation benefit. Neither evaluates whether a particular update helps
prediction beyond its source. Section~\ref{sec:settlement} evaluates that effect directly
(Appendices~\ref{app:decoder-controls} and~\ref{app:results-20260913}).

\FloatBarrier

\subsection{External validation measures whether an update transfers}
\label{sec:settlement}
Settlement checks an update before making it permanent. From current fast
weights $W$, it forms a temporary candidate $W+\delta$ and compares both states
on the same independent real-text passage $q$, without changing attention:
\begingroup
\setlength{\abovedisplayskip}{5pt}
\setlength{\belowdisplayskip}{5pt}
\setlength{\abovedisplayshortskip}{5pt}
\setlength{\belowdisplayshortskip}{5pt}
\begin{equation}
A(\delta;W,q)=L(q;W)-L(q;W+\delta).
\label{eq:advantage}
\end{equation}
\endgroup
Positive $A$ means that the candidate predicts $q$ better, so Settlement
commits $\delta$ only when $A\geq0$. Figure~\ref{fig:stage-story} locates large
costs accumulated over an eight-chunk passage; Settlement makes the finer,
candidate-level decision: an update must not increase loss on $q$.
Under this rule, 22/936 and 18/312
candidates are accepted at 125M and 760M.
Each candidate is evaluated on top of earlier accepted updates, because changes
that appear safe separately can be harmful when combined
(Appendix~\ref{app:commitment}). Settlement uses the next arriving real passage,
or retained prefix text when none arrives. \emph{The rule does not use the
candidate's source label}, but it requires identifiable real text for the
comparison. Algorithm~\ref{alg:settlement} in
Appendix~\ref{app:settlement-algorithm} gives the full procedure.

\begin{figure}[!ht]
\centering
\includegraphics[width=\linewidth]{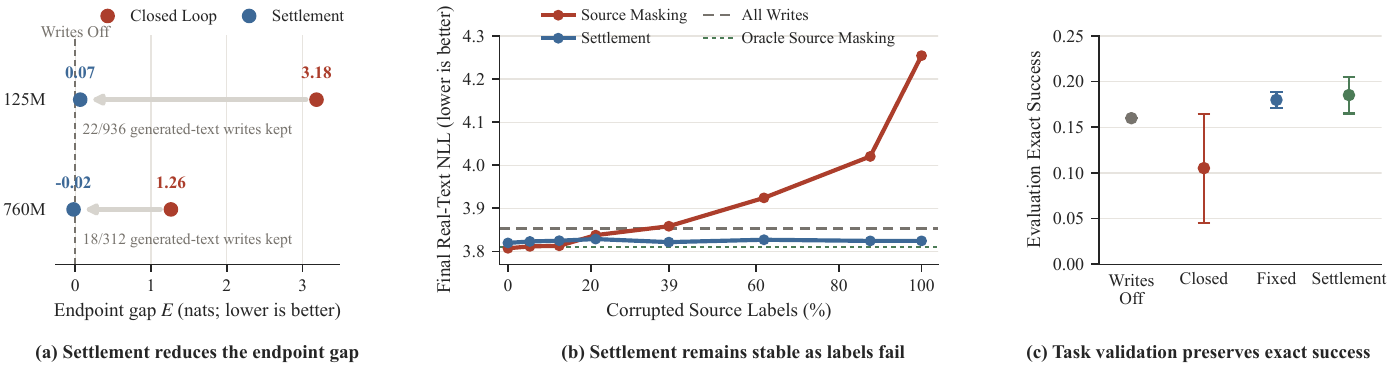}
\caption{\textbf{Independent evidence evaluates transfer.} (a) Settlement leaves
little endpoint residual. (b) Source Masking deteriorates under corrupted labels;
Settlement remains stable. Lower is better in (a--b). (c) Task-metric Settlement
uses WebShop reward to select updates; the plot shows exact success
(mean $\pm$ SD; five paired seeds; shared Writes Off). Higher is better. Details
are in Appendices~\ref{app:settlement-completed}--\ref{app:provenance-noise}
and~\ref{app:agent-credit}.}
\label{fig:settlement-provenance}
\end{figure}

Settlement leaves endpoint gaps of $.07$ nats at 125M and $-.02$ at 760M
(Figure~\ref{fig:settlement-provenance}a); both intervals include zero. The
improvement comes from validation rather than merely smaller updates or better
context. Without validation, quarter-sized writes, read-only real text, and
their combination leave $.90$, $1.04$, and $.28$ nats; adding Settlement puts
every variant within $.04$ nats of Writes Off
(Appendix~\ref{app:settlement-components}). On a matched all-real stream,
Settlement accepts 89 of 113 candidates. Relative to Reject All, it reduces
the first-to-last NLL change by $.0534$ nats $[.0299,.0721]$; its paired
advantage over Write All is $.0083$ $[.0020,.0161]$. Thus validation retains
measurable real-text learning (Appendix~\ref{app:settlement-pareto}).

\paragraph{When is external validation useful beyond Source Masking?}
Source Masking keeps real-text writes and rejects writes labeled as generated;
Settlement instead evaluates the proposed state. With correct labels, Source
Masking is simpler and has slightly lower NLL ($3.8072$ versus $3.8196$).
We then corrupt the observed source labels while leaving the underlying stream
unchanged. Source Masking begins to accept generated writes and reject real
ones, so its NLL rises. Settlement instead evaluates the candidate's prediction.
Its mean NLL is lower from $21.2\%$ observed error; paired intervals exclude
zero at the evaluated error rates of $38.9\%$ and above
(Figure~\ref{fig:settlement-provenance}b). Thus
Source Masking is simpler with reliable labels; validation is more robust to
label errors in this experiment. Full results appear
in Appendix~\ref{app:provenance-noise}.

Source Masking cannot choose among updates learned from an agent's own actions:
every candidate has the same ``generated'' label. Rejecting them all is Writes
Off; retaining them all is Closed Loop. WebShop \citep{webshop} provides this setting through
prompt--action updates with model-generated action targets. Settlement instead
checks validation reward, accepts 10 of 180 candidate blocks, and reaches
$.1853$ exact success, compared with $.1053$ for
Closed Loop and $.1600$ for the shared Writes Off reference; the adapting
policies use five paired seeds
(Figure~\ref{fig:settlement-provenance}c). The task metric therefore distinguishes
updates that their common source label cannot (Appendix~\ref{app:agent-credit}).
In ALFWorld household tasks \citep{alfworld}, two of three Closed Loop seeds
complete none of the 134 unseen tasks at each update scale. Settlement instead
achieves $88.1\%$--$94.0\%$ unseen-task success across seeds, retaining
$19\%$--$33\%$ of update blocks across the two scales
(Appendix~\ref{app:alfworld}).
\FloatBarrier

\section{Related Work}
\label{sec:scope}
\label{sec:implications}
\paragraph{Writable inference-time state.}
Fast weights provide memory that changes during sequence processing
\citep{schmidhuber1992fast,ba2016fast}. Linear attention interprets recurrent
state as a fast-weight matrix \citep{schlag2021linear}, while dynamic evaluation
updates parameters on recent context \citep{krause2018dynamic}. TTT layers,
TTT-E2E, Titans, and In-Place TTT bring learned update rules or writable memory
to language models \citep{tttlayers,tttE2E,titans,inplaceTTT}. We study the
case in which this state also generates its next training example.

\paragraph{Test-time adaptation and stability.}
Test-time training adapts on deployment inputs through self-supervised losses,
entropy minimization, or retrieved neighbors
\citep{sun2020ttt,tttplus,wang2021tent,hardt2024test}. Continual methods limit
erroneous pseudo-label accumulation through teacher averaging, restoration, or
reset \citep{arazo2020pseudo,cotta,niu2023towards,press2023rdumb}. Our setting
differs because the adapted model produces later inputs, a within-stream form
of performative prediction \citep{perdomo2020performative}. Fixed Generation
and Recorded Replay separate changes to future text, attention, and weights.

\paragraph{Generated-data feedback and update selection.}
Recursive synthetic-data studies show that training successive model
generations on generated samples can reduce quality or diversity, while mixing
in real data can slow or prevent collapse
\citep{selfconsuming,shumailov2024collapse,alemohammad2024mad,heatdeath,
gerstgrasser2024collapse}. Neural decoding has a related but distinct failure:
locally likely tokens can produce repetitive continuations, motivating nucleus
sampling and repetition-aware objectives
\citep{holtzman2020curious,welleck2020unlikelihood,xu2022breakloop}. Our object of
study is one adapting model in one stream, where weights change later text and
learning from that text changes the weights again. Decoder and replay controls
show why diversity and synthetic origin alone do not determine persistent cost.

Methods for continual learning constrain gradients or preserve performance on
stored examples \citep{lopezpaz2017gem,chaudhry2019agem}; validation-based
reweighting favors examples whose gradients improve held-out data
\citep{ren2018reweight}. Self-improvement methods likewise learn from generated
rationales, feedback, or successful trajectories
\citep{zelikman2022star,huang2023selfimprove,shinn2023reflexion}. More closely
related, aTTT studies within-episode feedback from online updates and
downweights repeated update tokens \citep{agenticttt}; VANE validates visual
updates on later observations \citep{vane}. Our controlled comparisons isolate
the feedback pathway in long text streams; Settlement evaluates the accumulated
fast-weight state on independent real text before retaining an update.

\section{Discussion and Conclusion}
\paragraph{What causes the failure?}
With a frozen generator, the learner updates on synthetic text yet stays near
Writes Off. Loss grows when adapted weights also generate later training
chunks: each update can change both the model and its future data. Late harm
is concentrated: four of 24 source sequences account for every fixed-receiver
write cost above $0.5$ nats. Affected source--receiver pairs remain above that
threshold at later sampled positions. The loop makes some trajectories
persistently harmful, not every late passage.

\paragraph{How does the causal decomposition support this explanation?}
Fixed Generation keeps the generated source and update count but breaks the
link from updates to later training text, removing $98.3\%$ and $98.8\%$ of
the matched Closed Loop gap at 125M and 760M. Recorded Replay then gives
identical receivers the same degraded tokens: reading raises real-text loss,
and retaining updates raises it further, including at 3B. Finally, one update
fits its generated chunk better but predicts the next real passage worse; that
cost grows after Closed Loop adaptation. The controls locate the failure in
feedback and poor transfer, not synthetic origin alone. In WebShop, breaking
feedback likewise raises mean exact success from $.1053$ to $.1800$ across
five paired seeds.

\paragraph{What reduces the damage?}
Real text works best when it repeatedly interrupts self-writing: with $31\%$
real chunks, even spacing leaves $.0694$ nats of excess harm versus $.7155$
for bursts. Decoder changes reduce harm without identifying which update
helps. Settlement directly compares current and proposed states on independent
real text before commitment. In separate experiments, it nearly matches
Writes Off on generated streams and retains useful updates on real-text
streams. Source Masking is simpler with reliable labels; at tested error rates
of $38.9\%$ and above, Settlement has lower NLL when real validation text is
available. The criterion is whether the complete state to be retained,
including combined parallel updates, improves performance beyond its source.

\label{endofmain}
\clearpage
\bibliography{refs}
\bibliographystyle{style/iclr2027_conference}

\medskip
\appendix
\section{Experimental scope and implementation}
\label{app:claims}
\label{app:implementation}
The experiments answer three questions: whether persistent self-writing causes
long-horizon damage, which part of the feedback loop causes it, and whether
independent evidence can screen a proposed update. Every conclusion comes from
a treatment and control with the shared settings listed below; online generation
may produce different text under different policies. Unless stated otherwise, language-model intervals are
95\% bootstrap intervals over books after averaging seeds within each book.

\paragraph{Stable book identities.}
The cross-scale, Fixed Generation, and decoder experiments use the same six
PG-19 documents, denoted \emph{canonical books 2--7}, with five seeds and a
128K horizon. Exposure and validation require longer uninterrupted documents,
so each uses one fixed long-book set shared by every policy in that comparison.
The manifests record document hashes, selection thresholds, and scripts.

\paragraph{Protocol map.}
\label{app:protocol-map}
Table~\ref{tab:protocol-map} maps each paper claim to its shared configuration
and controlled comparison. Long-horizon $H$ and $E$ use Writes Off as the reference;
local comparisons hold the received text fixed and compare retaining versus discarding its updates.
$H$ compares first-to-last changes; $E$ compares final endpoints and is used
for intervention policies. They coincide when the initial losses match.

\begin{table}[!ht]
\caption{Main experiment configurations and why they differ. Each row compares
policies under one shared configuration; books are the independent units for
language-model intervals.}
\label{tab:protocol-map}
\centering\small\setlength{\tabcolsep}{2.5pt}\renewcommand{\arraystretch}{.96}
\begin{tabular}{p{.20\linewidth}p{.42\linewidth}p{.28\linewidth}}
\toprule
Question & Shared configuration & Controlled comparison\\
\midrule
Long horizon & 125M, 760M, 3B; canonical six books; five seeds; 128K; width 8 & Closed Loop versus Writes Off\\
Feedback path & 125M and 760M; canonical six books; five seeds; 128K; Fixed Generation added & Closed Loop versus Fixed Generation\\
Read vs. write & Fixed recorded tokens; 125M: eight receiver books, five seeds; 3B: six source and six disjoint receiver books & Read Only versus Read + Write\\
Single update & Eight books and four stream positions; identical state and source text in each branch & Keep versus discard one update\\
Decoder/exposure & Decoder: canonical setup; exposure: one eight-book long-stream set, five seeds & Decoder endpoints versus a common Writes Off reference; evenly spaced versus bursty real text\\
Validation & Scale-specific long-book sets; each scale has its own paired Writes Off trajectory & Closed Loop versus Settlement\\
Agents (WebShop) & Qwen3-4B LoRA; five paired seeds; fixed test-goal evaluation & Closed Loop, Fixed Generation, Settlement, Writes Off\\
Agents (ALFWorld) & Qwen3.8-27B LoRA; three paired 274-task streams; prequential scoring; update scales 8 and 16 & Closed Loop, Settlement, Writes Off\\
\bottomrule
\end{tabular}

\end{table}

\paragraph{Model and training.}
The 125M TTT-E2E model has 12 layers and width 768, with an 8192-token sliding
attention window in the first nine layers and a SwiGLU fast-weight
MLP in each of the final three layers. Each sequence has its own fast weights.
One clipped inner-loop SGD step follows each 1024-token chunk. Unless a
decoder control is specified, temperature is one and top-$p$ is $.95$.
Training follows
the official 4800-step recipe on 2.5B DCLM tokens \citep{dclm}, followed by the
32K length-extension schedule on PG-19 \citep{pg19} in place of Books.
The experiment manifests record complete SHA-256 digests for the 125M model
file and the PG-19 validation array.

\paragraph{Conversion and numerical checks.}
The PyTorch and official-format models have identical parameter counts.
Conversion changes the sequence loss by $3.6\times10^{-5}$ over eight inner-loop
chunks; every chunk loss differs by less than $10^{-4}$. A rematerialized
inner-loop backward agrees with a naive second-order implementation to
$6\times10^{-16}$ in fp64. For the released 3B/128K model, the converted
training forward changes sequence loss by $1.05\times10^{-4}$ and gives
token-level NLL correlation $0.9996$. This verifies the converted training
forward, not exact parity of the 3B streaming decoder with the official decoder.

\paragraph{Streaming evaluation.}
The standard 128K schedule starts with eight real-text chunks, followed by
generated chunks and periodic real-text evaluations. The original schedule has
15 branch evaluations and 105 generated chunks; the new cross-scale suite reports
16 NLL values including its initial reference. Evaluation never updates weights.
Evaluations snapshot and restore the complete carried state, so evaluation text does
not become later generation context. The state contains fast weights, attention
caches, positions, held updates, and partial-chunk buffers. We distinguish a
condition's first-to-last NLL change $D_c$ from the paired excess
$H_c=D_c-D_{\mathrm{off}}$. When initial losses are matched exactly, this
equals the final-evaluation difference. Other tables explicitly use the endpoint
gap $E_c=L_{c,\mathrm{last}}-L_{\mathrm{off},\mathrm{last}}$.

\paragraph{Write strength, retention, and state restoration.}
The write-strength multiplier scales the clipped update, not the loss
before clipping. For $w\in\{1/8,1/4,1/2,1,2,4\}$, a single-step norm obeys
$\|\Delta W(w)\|/(w\|\Delta W(1)\|)=1$ to a worst relative error of
$1.15\times10^{-7}$; $w=0$ gives exactly zero update. Loss scaling before a
saturated norm clip would not implement the same controlled operation. Over 12 chunks,
net drift is not linear in $w$ because update directions partly cancel.
Retention applies $W\leftarrow W_0+\lambda(W-W_0)$ once per chunk, including
chunks whose proposals are rejected. With no accepted updates, six chunks at
$\lambda=.9$ give exactly the expected displacement factor $.531441$.
Snapshot--restore returns all 39 carried tensors bitwise identical and four
scalars equal. A teacher-forced 12-chunk regression check with interleaved
evaluations matches the losses obtained without evaluation exactly. These
checks do not establish bitwise reproducibility of all stochastic generation
trajectories.

\subsection{Generator isolation and chunk boundaries}
\label{app:control-audit}
The Fixed Generation control uses a separate cache advanced only by
the frozen generator. On two books, perturbing
receiver weights, KV, positions, and buffers leaves this generator unchanged.
Keyed per-row sampling is invariant to batch reordering, and a writes-off
null comparison is bitwise identical.

The canonical implementation starts each generated chunk with a real boundary
token, and every paired policy uses the same rule. Teacher-forcing recorded
tokens is not bitwise equivalent to online decoding; the comparison gives a maximum
logit difference $.1875$ and mean KL $.000287$.

\subsection{Feedback and displacement control}
\label{app:audited-drift}
This suite uses the 125M model, eight receiver books, five seeds, 128
chunks, and the canonical boundary rule. All four policies use the same code
path. Real-text evaluations snapshot and restore
state. The analysis first averages seeds within book, then uses a 20,000-draw
paired bootstrap over the eight books. CUDA sampling and the disabled
token-level entropy observer are recorded in the experiment manifest.

\begin{table}[ht]
\caption{\textbf{Matched 125M feedback and displacement control.} Drift is
$\|W_t-W_0\|/\|W_0\|$. Real-Text Learning contains no generated tail, so token-diversity
statistics are not applicable.}
\label{tab:audited-drift}
\centering\small
\begin{tabular}{lrrrr}
\toprule
Policy & Final clean NLL & Final relative drift & Tail Distinct-2 & Tail Repeated-4\\
\midrule
Closed Loop & $6.5084$ & $.00857$ & $.0084$ & $.9902$\\
Writes Off & $3.6776$ & $.00500$ & $.8493$ & $.0217$\\
Fixed Generation & $3.7483$ & $.01387$ & $.8779$ & $.0082$\\
Real-Text Learning & $3.3232$ & $.01759$ & --- & ---\\
\bottomrule
\end{tabular}

\end{table}

Paired endpoint contrasts are: Closed Loop minus Writes Off $2.8308$
$[1.8171,3.8598]$; Fixed Generation minus Writes Off $.0707$
$[.0510,.0924]$; Real-Text Learning minus Writes Off $-.3544$
$[-.6048,-.1236]$; and Closed Loop minus Fixed Generation $2.7601$
$[1.7555,3.7881]$. Fixed Generation and Real-Text Learning move farther from initialization
than Closed Loop while causing far less harm. Parameter displacement is therefore
not a sufficient explanation of the closed-loop loss. The independent
Fixed Generation suite in Appendix~\ref{app:new-scale} remains a
replication under a separate protocol.

\subsection{Model-specific contamination screening}
\label{app:contamination}
The first two candidate books have anomalously low loss under the larger
released models. We compare unadapted first-chunk loss against
the 125M reproduction. The larger ratios on books 0--1 are consistent with
model-specific familiarity, but cannot establish the exact training
membership of any book. Table~\ref{tab:contamination-curated} defines the ratio
explicitly. The screened 3B set uses canonical books 2--7. Screening removes a visible
anomaly; it does not prove absence of all contamination. The 760M and 3B
training lineages also differ, so their damage magnitudes do not establish
a scaling law.

\begin{table}[ht]
\caption{Unadapted loss ratio $L_{125M}/L_{\mathrm{model}}$ on the
same book. A large ratio means unusually low loss under the larger model.}
\label{tab:contamination-curated}
\centering\small
\begin{tabular}{lrrrrrrrr}
\toprule
Model & 0 & 1 & 2 & 3 & 4 & 5 & 6 & 7\\
\midrule
760M, DCLM & 3.69 & 1.44 & 1.15 & 1.23 & 1.18 & 1.19 & 1.14 & 1.18\\
3B, DCLM & 10.35 & 4.12 & 1.30 & 1.33 & 1.29 & 1.32 & 1.30 & 1.28\\
3B, Books finetune & 8.83 & 2.67 & 1.44 & 1.69 & 1.44 & 1.41 & 1.37 & 1.46\\
\bottomrule
\end{tabular}

\end{table}

\FloatBarrier

\section{Replication and boundaries of the failure regime}
\label{app:new-scale}
This appendix follows Section~\ref{sec:failure}. It first confirms Fixed
Generation at a second model scale, then asks which properties
of the stream change the severity of the failure. The evidence gives two clear
boundaries: decoder choice changes both when harm becomes visible and how fast
it grows, whereas regular real-text writes sharply suppress it.

\subsection{Fixed Generation confirmation at 125M and 760M}
The three-condition suite compares Closed Loop, Writes Off, and Fixed
Generation. Table~\ref{tab:new-scale-curated} evaluates both scales on the
same six canonical books with five seeds per book. Initial evaluations agree across policies
within each scale, so the endpoint contrasts are directly comparable within a
row and use the same first-to-last definition as Table~\ref{tab:current-results}.

\begin{table}[ht]
\caption{Fixed Generation confirmation on the same six receiver
books. Intervals resample books after averaging seeds.}
\label{tab:new-scale-curated}
\centering\small
\begin{tabular}{llrrr}
\toprule
Model & Condition & Initial NLL & Final NLL & Final excess over Writes Off [95\% CI]\\
\midrule
125M & Writes Off & 3.3469 & 3.3845 & 0\\
 & Fixed Generation & 3.3469 & 3.4361 & .0516 [.0248,.0772]\\
 & Closed Loop & 3.3469 & 6.3942 & 3.0097 [1.8575,4.1972]\\
\midrule
760M & Writes Off & 2.7913 & 2.9383 & 0\\
 & Fixed Generation & 2.7913 & 3.0116 & .0733 [.0538,.0939]\\
 & Closed Loop & 2.7913 & 8.9405 & 6.0021 [4.2695,7.7347]\\
\bottomrule
\end{tabular}

\end{table}
At both scales, fixing the generator leaves only a small fraction of the Closed
Loop endpoint gap. The independent 3B confirmation in
Appendix~\ref{app:replay-3b-confirmation} instead holds the received text fixed
and measures the additional cost of retaining its updates.

\subsection{Decoder and reset controls change the rate of damage}
\label{app:decoder-controls}
\label{app:sampling-support}
Six decoder policies and two reset schedules use the same 125M
model, canonical books 2--7, five seeds, and a 128K schedule with 105 generated
chunks. All rows use the same book-blocked estimator as
Table~\ref{tab:current-results}. The repetition penalty is applied to tokens
already generated within the current chunk; typical-$p=.95$ replaces rather
than composes with top-$p$ sampling. A reset restores fast weights to their
initial values every $K$ chunks but leaves the attention cache intact.

\begin{table}[H]
\caption{Canonical decoder and reset controls. The endpoint gap uses the same
canonical Writes Off endpoint for every row; late slope is the mean change per evaluation over the
final four intervals. Lower is better in both columns.}
\label{tab:decoder-controls}
\centering\scriptsize\setlength{\tabcolsep}{2.8pt}
\begin{tabular}{lrr}
\toprule
Control & \shortstack{Endpoint gap to\\canonical Writes Off} & \shortstack{Late slope\\(nats/evaluation)}\\
\midrule
top-$p=.95$ & $3.0097$ & $+.1339$\\
top-$p=.99$ & $3.3531$ & $+.1497$\\
typical-$p=.95$ & $2.9292$ & $+.0948$\\
full support & $1.4520$ & $+.1979$\\
$T=.7$, top-$p=.95$ & $2.2246$ & $+.1704$\\
repetition penalty $1.1$ & $\mathbf{.8794}$ & $\mathbf{+.0027}$\\
reset every 8 chunks & $\mathbf{.1749}$ & $\mathbf{+.0056}$\\
reset every 32 chunks & $1.9218$ & $+.0623$\\
\bottomrule
\end{tabular}
\end{table}
Endpoint and late slope answer different questions. Full-support sampling has
a smaller 128K endpoint than the default decoder, but its late slope is the
largest in the table: it delays the visible failure without flattening the
curve. Fixed Generation, reset every eight chunks, and the repetition penalty
all have late slopes below $.006$; the remaining continuing controls have late
slopes above $.062$. Reset frequency matters sharply. Resetting every eight
chunks leaves an endpoint gap of $.1749$ $[.084,.321]$ (5.8\% of the canonical Closed Loop
damage), whereas resetting every 32 leaves a gap of $1.9218$
$[.949,3.158]$ (63.9\%). Thus a short reset
window is an effective safety baseline, but it also limits how long adaptation
can persist.

We measure that trade-off on real-text streams. Ordinary persistent writing
improves NLL by $.0850$ $[.0444,.1260]$ over Writes Off. Resetting every eight
chunks preserves a $.0375$ $[.0195,.0526]$ benefit, or 44.1\% of the persistent
adaptation gain. Reset therefore controls runaway accumulation by shortening
the lifetime of every update, including useful ones. The repetition penalty
also changes the entire future stream; neither control decides whether a
particular fixed passage should be written.

For intuition, consider an idealized SGD step at a fixed context. Sampling
from $q_W(x)\propto p_W(x)^{1/T}$ on fixed support $A$ and applying
$\eta\nabla_W\log p_W(x)$ gives expected increment
$\eta T\nabla_W\log\sum_{x\in A}p_W(x)^{1/T}$. At $T=1$ with full support,
this mean is zero; truncation instead increases retained probability mass to
first order. The derivation assumes an infinitesimal unclipped step and a fixed
context and therefore does not predict a zero finite-horizon effect. The paired
experiment above supplies the relevant empirical result: truncation amplifies
damage but is not required for it.

\subsection{External-text exposure bounds the failure regime}
\label{app:exposure-density}
We vary the number and placement of real-text write slots while keeping the
125M model, the dedicated long-book set $[0,1,27,28,29,43,44,46]$, five
seeds, 8K real prefix, 105 live slots,
temperature one, and top-$p=.95$ fixed. \emph{All Writes} retains updates from
both real and generated slots. \emph{Writes Off} retains real-text updates but
discards generated-text writes while still reading those tokens. Thus $H$ is
the excess NLL change caused by retaining generated-text writes at each exposure
density.
The exposure sweep needs longer uninterrupted documents than the canonical
comparison, so every exposure policy instead uses the same eight long books.
The main-text panel reports $H/H_{0\%}$ within this set; the table below gives
the corresponding absolute values.

\begin{table}[ht]
\caption{Real-text cadence controls excess NLL change. $H$ is the excess NLL change
of retaining generated writes; $B$ is the benefit of retaining real-text
writes. Intervals use a paired book bootstrap after averaging seeds.}
\label{tab:exposure-density}
\centering\small\setlength{\tabcolsep}{4pt}
\begin{tabular}{rrrrrr}
\toprule
Real fraction & Schedule & Real slots & Max self burst & $H$ [95\% CI] & $B$ [95\% CI]\\
\midrule
0\% & Even & 0 & 105 & $1.1858;[.5326,1.9430]$ & $.0000;[.0000,.0000]$\\
5\% & Even & 5 & 17 & $1.0275;[.5679,1.6195]$ & $.1350;[.0557,.2578]$\\
10\% & Even & 11 & 8 & $.4389;[.2517,.6465]$ & $.0933;[.0729,.1155]$\\
20\% & Even & 21 & 4 & $.0806;[.0657,.0981]$ & $.0802;[.0585,.1052]$\\
31\% & Even & 33 & 3 & $.0694;[.0534,.0843]$ & $.1051;[.0668,.1431]$\\
31\% & Bursty & 33 & 12 & $.7155;[.3730,1.1578]$ & $.1040;[.0502,.1527]$\\
\bottomrule
\end{tabular}

\end{table}
Increasing evenly spaced real-text exposure reduces excess NLL change while
preserving a positive real-text adaptation benefit. Density is not the whole
story: at 31\%, grouping the same 33 real slots together increases the longest
uninterrupted stretch of self-generated chunks from 3 to 12 and increases $H$
by an order of magnitude. The controlled quantity is therefore real-text
cadence: both how much real text arrives and the longest interval without it.

\subsection{Teacher-forced real-text payoff}
\label{app:payoff-final}
The matched 125M payoff experiment uses eleven fixed teacher-forced real-text
books. Benefit is the loss under \emph{No Updates} minus the loss under the
evaluated writing policy; positive values mean that adaptation helps. Ordinary
sequential writing gives $.0340$ nats of benefit. Because the text is fixed,
the improvement cannot come from changing future training tokens. The same
suite evaluates whether validation preserves this gain, including the failure
of individually validated but jointly committed updates; the full comparison
appears in Table~\ref{tab:commitment-appendix}.

\FloatBarrier

\section{Causal decomposition: feedback, attention, and persistent weights}
\label{app:causal-extra}
This appendix follows the causal order of Section~\ref{sec:causal}. We first
separate the generator from the learner, then replay identical text with and
without updates, and finally branch before a single update. Each step removes
one source of variation, moving from the full online loop to a paired local
comparison.
\subsection{Fixed Generation and Recorded Replay}
The separate eight-book displacement suite is reported in
Appendix~\ref{app:audited-drift}: Closed Loop minus Writes Off is $2.8308$
$[1.8171,3.8598]$, while Fixed Generation minus Writes Off is $.0707$
$[.0510,.0924]$. The matched 760M result is reported in
Appendix~\ref{app:new-scale}. These controls show that continuing to update on
generated text need not reproduce the large closed-loop loss. They change
subsequent text as well as its dependence on adapting weights, so they do not
hold content fixed.

The Recorded Replay comparison instead supplies the same recorded tokens to the receiver
with writing enabled or disabled. Relative to the suite's Writes Off control, Recorded Replay without
writes gives $1.606$ nats of excess and replay with writes gives $3.874$,
an additional $2.268$ nats. This paired comparison isolates the contribution
of writing that replayed text. Comparing live closed generation with replay
does not isolate an additional feedback effect, because they contain different
text. The 3B experiment below is the large-model confirmation.

\subsection{3B fixed-text replay confirmation}
\label{app:replay-3b-confirmation}
This experiment evaluates whether retaining updates from a fixed recorded
stream adds cost beyond reading exactly the same tokens. Six source books are
disjoint from six receiver books. Two cyclic source--receiver
mappings and five independently sampled recordings supply the matched
read/write comparisons. For each pair,
\[
J=(L_{\rm write,last}-L_{\rm write,first})
 -(L_{\rm read,last}-L_{\rm read,first}).
\]
The paired trajectories have exactly equal initial NLL and identical received
token hashes. A 20,000-draw two-way source/receiver cluster bootstrap gives
$J=0.1927$ $[0.1043,0.3272]$; clustering only by source gives $0.1927$
$[0.1181,0.2941]$. Both analyses account for shared source books rather than
treating trajectories as independent samples.

\begin{table}[ht]
\caption{\textbf{The 3B passage-level write cost is positive across every source
and receiver aggregate.} The statistical interval is clustered by source and
receiver, not by individual trajectory.}
\label{tab:replay-3b-confirmation}
\centering
\begin{minipage}[t]{.47\linewidth}
\centering\small\setlength{\tabcolsep}{5pt}
\begin{tabular}{crr}
\toprule
Source book & Mean $J$ & Leave-one-out $J$\\
\midrule
12 & .1862 & .1940\\
13 & .1953 & .1922\\
14 & .0617 & .2189\\
15 & .1331 & .2046\\
16 & .1602 & .1992\\
17 & .4196 & .1473\\
\bottomrule
\end{tabular}
\end{minipage}\hfill
\begin{minipage}[t]{.47\linewidth}
\centering\small\setlength{\tabcolsep}{8pt}
\begin{tabular}{cr}
\toprule
Receiver book & Mean $J$\\
\midrule
2 & .2872\\
3 & .1824\\
4 & .1302\\
5 & .1042\\
6 & .1343\\
7 & .3178\\
\bottomrule
\end{tabular}
\end{minipage}

\end{table}

\subsection{Paired one-update comparison}
The 125M experiment branches at generated positions 1, 33, 65, and 97 from
Closed Loop, Writes Off, or Fixed Generation histories. The branches share the starting
state and current text; one retains that text's update and the other skips it.
Four draws of four future chunks use paired random numbers. All 15 history
trajectories and 60 paired configurations completed, giving 480 book-level
observations. We average seeds and positions within each book before a
20,000-draw paired book bootstrap. Each row in
Table~\ref{tab:single-update-detail} is a paired keep-minus-discard comparison
within its starting states; differences between rows also change the text and
history, and do not isolate receiver susceptibility alone.

\begin{table}[ht]
\caption{Keep minus discard for one update: negative source NLL improves
fitting; positive real NLL harms independent prediction. Intervals are
95\% book-bootstrap intervals.}
\label{tab:single-update-detail}
\centering\small\setlength{\tabcolsep}{3pt}
\begin{tabular}{lccc}
\toprule
History & $\Delta$ source NLL & $\Delta$ real NLL & $\Delta$ future repeated-4\\
\midrule
Closed Loop & $-.0354\;[-.0384,-.0317]$ & $.0381\;[.0098,.0718]$ & $.0086\;[.0020,.0184]$\\
Writes Off & $-.1374\;[-.1453,-.1285]$ & $.0045\;[.0037,.0053]$ & $.0049\;[-.0079,.0248]$\\
Fixed Generation & $-.1449\;[-.1482,-.1407]$ & $.0018\;[.0014,.0023]$ & $.0017\;[-.0034,.0063]$\\
\bottomrule
\end{tabular}

\end{table}

\begin{table}[ht]
\caption{Mean one-update real-text NLL effects by position, 125M.
Positions differ in both current text and receiver state.}
\label{tab:single-position-curated}
\centering\small
\begin{tabular}{lrrrr}
\toprule
History & Position 1 & 33 & 65 & 97\\
\midrule
Closed Loop & .0062 & .0087 & .0250 & .1126\\
Writes Off & .0063 & .0044 & .0038 & .0037\\
Fixed Generation & .0042 & .0008 & .0011 & .0012\\
\bottomrule
\end{tabular}

\end{table}
The paired comparison uses matched random numbers and measures update norms
from the applied parameter increment. Reimplementation checks reproduce the
reported aggregate outcomes; the analysis concerns paired outcome differences
rather than bitwise execution traces.

\subsection{Gradient conflict and one-update transfer cost}
\label{app:gradient-conflict}
\paragraph{Purpose.}
The paired experiment in Section~\ref{sec:single} shows that retaining one
generated-text update can worsen prediction of new real text. Here we ask
whether that harm is related to a conflict already present before the update:
does the generated chunk ask the weights to move in a direction that opposes
what the next real passage needs?

\paragraph{Quantities computed for each candidate.}
Let $W$ be the shared weights before branching, $x$ the generated chunk, and
$q$ the next independent real passage. We compute both gradients at $W$:
\[
g_{\rm self}=\nabla_W L(x;W), \qquad
g_{\rm real}=\nabla_W L(q;W).
\]
The optimizer uses $x$ to produce the actual candidate change $\Delta W$,
including the same clipping and normalization used in the main experiment.
The Keep branch applies $W+\Delta W$; the Discard branch remains at $W$. With
attention held fixed, their difference on $q$ is
\[
\Delta L_{\rm real}=L(q;W+\Delta W)-L(q;W).
\]
This is the measured one-update cost. Positive $\Delta L_{\rm real}$ means that
retaining the update made the real passage harder to predict. We also report
$\Delta L_{\rm self}=L(x;W+\Delta W)-L(x;W)$; a negative value confirms that
the same update improved its generated source.

\paragraph{Two pre-update signals.}
The first signal is gradient cosine,
\[
s_{\rm cos}=
\frac{g_{\rm self}^{\top}g_{\rm real}}
{\lVert g_{\rm self}\rVert\lVert g_{\rm real}\rVert}.
\]
When $s_{\rm cos}<0$, the two gradients point in opposing directions. A step
that reduces the generated-chunk loss then tends to increase real-text loss.
The second signal uses the actual proposed change,
\[
s_{\rm first}=g_{\rm real}^{\top}\Delta W
\approx \Delta L_{\rm real}.
\]
It is the first-order prediction of the real-text loss change: larger positive
values predict greater harm.

\paragraph{Correlation analysis.}
For each model and preceding history, eight books, three seeds, and four stream
positions give $8\times3\times4=96$ candidates. We compute Spearman correlation
across these candidates. If conflict predicts harm, $s_{\rm cos}$ should have
negative correlation with $\Delta L_{\rm real}$, whereas $s_{\rm first}$ should
have positive correlation. Spearman correlation evaluates whether each signal
ranks the more harmful candidates; its value is not an NLL increase. The
reported intervals use book-level resampling.

\begin{table}[ht]
\caption{Gradient conflict predicts the measured cost of retaining one
generated-text update. $\Delta L_{\rm real}$ is keep minus discard on the next
real passage; negative cosine means the source and real-text gradients oppose
one another. Negative mean $\Delta L_{\rm self}$ indicates improved source fit
on average.}
\label{tab:gradient-conflict-detail}
\centering\scriptsize\setlength{\tabcolsep}{3pt}
\begin{tabular}{lrrrr}
\toprule
Model / history & $\rho(\cos,\Delta L_{\rm real})$ &
$\rho(g_{\rm real}^{\top}\Delta W,\Delta L_{\rm real})$ &
$\Delta L_{\rm self}$ & Mean $\cos$\\
\midrule
125M / Closed Loop & $-.502\;[-.642,-.305]$ & $.704\;[.521,.829]$ & $-.033$ & $-.093$\\
125M / Writes Off & $-.788\;[-.826,-.720]$ & $.794\;[.721,.838]$ & $-.136$ & $-.005$\\
760M / Closed Loop & $-.633\;[-.808,-.383]$ & $.632\;[.449,.781]$ & $-.051$ & $-.103$\\
760M / Writes Off & $-.731\;[-.810,-.666]$ & $.751\;[.685,.828]$ & $-.166$ & $-.004$\\
\bottomrule
\end{tabular}

\end{table}

\paragraph{Results.}
All four cosine correlations are negative ($-.502$ to $-.788$), and all four
first-order correlations are positive ($.632$ to $.794$); all eight confidence
intervals exclude zero. Thus, in both models and both preceding histories, the
candidate updates with stronger generated--real conflict cause larger measured
real-text costs.

The mean cosine adds a state-level observation. Generated and real gradients
are nearly orthogonal after Writes Off history ($-.005$ at 125M and $-.004$ at
760M), but become more opposing after Closed Loop history ($-.093$ and
$-.103$). This is not a generic property of any two gradients: two real-text
gradients have positive mean cosine in the same states ($.125/.481$ at 125M and
$.097/.481$ at 760M, Writes Off/Closed Loop). The analysis therefore identifies
a specific conflict between fitting generated text and transferring to real
text. It explains which single updates are harmful; Settlement instead evaluates
the complete accumulated state before committing it.

\subsection{In-place Adam: dose and history are separate considerations}
\label{app:inplace}
Qwen3-4B freezes all parameters except the final four
\texttt{mlp.down\_proj} matrices and takes one Adam step per chunk, with
gradient norm clipped at one. It uses no outer meta-learning or added modules.
Eight books, three seeds, batches of four, eight real warm-up chunks, and
128 total chunks supply Table~\ref{tab:inplace-adam}. The latest 8192 tokens
are re-encoded each chunk; naive KV truncation raised evaluation loss from about
3.04 to 8.85, while rebuilding restored it to 3.02. This numerical check motivates the
reported window handling. The comparison isolates ordinary Adam updates to
existing model parameters rather than an added TTT module.

\begin{table}[ht]
\caption{In-place Adam long-horizon control. The larger-rate excess is
$1.231$ [.580,2.350]; the smaller-rate excess is $.026$ [-.071,.122].}
\label{tab:inplace-adam}
\centering\small
\begin{tabular}{lrrrr}
\toprule
Condition & NLL change $D$ & Relative drift & Distinct-2 & Repeated-4\\
\midrule
Writes Off & -.060 & .00000 & .140 & .831\\
Adam $10^{-5}$ & -.034 & .00627 & .187 & .775\\
Adam $10^{-4}$ & 1.171 & .02399 & .008 & .991\\
\bottomrule
\end{tabular}

\end{table}
To verify that the larger learning rate is not simply unusable, we also adapt
the same parameter subset on teacher-forced real text. The future tokens are
fixed and therefore cannot be changed by the update. Positive
$B=L_{\rm no\ update,last}-L_{\rm real\ update,last}$ means that writing
improves prediction on independent real text.

\begin{table}[ht]
\caption{\textbf{The Qwen3-4B update configuration also learns useful real
text.} Two independently adapted states each contain four books; the bracket
is their range, not a book-level confidence interval.}
\label{tab:qwen-real-benefit}
\centering\small
\begin{tabular}{lrrr}
\toprule
Adam LR & $B$ [range of two state means] & State 0 & State 1\\
\midrule
$10^{-5}$ & $.1392\;[.1150,.1634]$ & $.1634$ & $.1150$\\
$10^{-4}$ & $.1680\;[.1317,.2043]$ & $.2043$ & $.1317$\\
\bottomrule
\end{tabular}

\end{table}
Both learning rates improve independent real-text prediction in both adapted states.
Together with Table~\ref{tab:inplace-adam}, this shows that the $10^{-4}$
configuration can help on real text while harming prediction when its future
training stream is self-generated.

\section{From heterogeneous harm to partial repairs}
\label{app:stage-sweep}
Section~\ref{sec:control} shows that late damage is heavy-tailed rather than
uniform. This appendix first gives the underlying distribution, then evaluates
three natural responses: weighting repeated tokens less, shrinking every
write, and inserting real text without writing it. Each response reduces harm,
but none directly determines whether a particular update transfers.

\subsection{A minority of source passages drives the late mean}
The fixed-receiver sweep replays eight-chunk passages from 24 source rows into
three receiver offsets at seven stream positions. For each passage, write
damage is subsequent real-text NLL after read-and-write minus NLL after
read-only processing from the same receiver state. Receiver offsets show
whether a source effect transfers across books; they do not create additional
independent source trajectories.

\begin{table}[ht]
\caption{Fixed-receiver stage replay. The increasing mean is produced by a
growing right tail; the median remains near zero.}
\label{tab:stage-curated}
\centering\small
\begin{tabular}{rrrrr}
\toprule
Source stage & Median & Mean & Standard deviation & Cases above .5 nats\\
\midrule
1 & .0422 & .0439 & .025 & 0/72\\
17 & .0040 & .0898 & .385 & 3/72\\
33 & .0039 & .1399 & .620 & 3/72\\
49 & .0042 & .3135 & 1.009 & 6/72\\
65 & .0047 & .4855 & 1.274 & 9/72\\
81 & .0105 & .4858 & 1.267 & 9/72\\
97 & .0064 & .7512 & 1.731 & 12/72\\
\bottomrule
\end{tabular}

\end{table}
All cases above $.5$ nats come from four of the 24 source rows. The same rows
produce large effects at all three receiver offsets. There are
$24\times3\times7=504$ paired costs in total. After first exceeding $.5$ nats,
those source--receiver pairs remain above it at later sampled positions.
Receivers are reset between comparisons; this measures the harm of later
source passages, not a receiver's recovery. A separate repetition-bin
comparison gives an unresolved early-to-late change of $.0587$
$[-.0382,.3286]$. Stream position and repetition therefore indicate risk but
do not identify every harmful passage.

\subsection{Repetition weighting mitigates but does not remove damage}
\label{app:results-20260913}
The expanded policy suite uses sixteen books and five seeds. Random and uniform
controls reduce update weight without using token identity; the comparison asks
whether repetition-aware weighting helps beyond simply learning less.

\begin{table}[ht]
\caption{Repetition weighting and real context reduce the endpoint gap but do
not reach Writes Off. Lower $E$ is better.}
\label{tab:attt-expanded}
\centering\small
\begin{tabular}{lrrr}
\toprule
Policy & Final NLL & $E$ [95\% interval] & Distinct-2\\
\midrule
Writes Off & 3.5767 & 0 & .8425\\
Closed Loop & 6.4620 & 2.8853 [2.0072,3.8310] & .0082\\
Repetition weighting & 5.1404 & 1.5637 [1.1858,1.9825] & .0469\\
Random dose & 5.6325 & 2.0558 [1.6339,2.4946] & .0086\\
Uniform dose & 6.0628 & 2.4861 [1.9510,3.0939] & .0155\\
Read-only real context & 4.9746 & 1.3979 [1.0988,1.7140] & .0730\\
\bottomrule
\end{tabular}

\end{table}
Repetition weighting improves on random and uniform dose assignment by $.4921$
and $.9224$ nats, respectively, while leaving $1.5637$ nats above Writes Off.
The result supports repetition as a useful symptom, not a sufficient admission
rule. Because these are online policies, later generated text also differs.

\subsection{Smaller writes expose a stability--adaptation trade-off}
We scale each post-clipping parameter increment by
$\alpha\in\{0,1/16,1/8,1/4,1/2,1,2\}$. The same scale is applied to
teacher-forced real-text writes. Generated-stream harm $H$ and real-text
benefit $B$ are both measured relative to $\alpha=0$.

\begin{table}[ht]
\caption{Uniformly smaller writes reduce both harm and useful adaptation.
Every evaluated nonzero dose has positive values for both quantities.}
\label{tab:dose-frontier}
\centering\small\setlength{\tabcolsep}{7pt}
\begin{tabular}{rrr}
\toprule
$\alpha$ & Excess NLL change $H$ [95\% CI] & Real-text benefit $B$ [95\% CI]\\
\midrule
$0$ & $.0000;[.0000,.0000]$ & $.0000;[.0000,.0000]$\\
$1/16$ & $.3856;[.1737,.6432]$ & $.0412;[.0249,.0579]$\\
$1/8$ & $.8583;[.5480,1.2276]$ & $.0543;[.0328,.0758]$\\
$1/4$ & $.9248;[.6525,1.3374]$ & $.0682;[.0413,.0942]$\\
$1/2$ & $1.0319;[.6424,1.4540]$ & $.0812;[.0507,.1106]$\\
$1$ & $1.1867;[.8512,1.5056]$ & $.0922;[.0579,.1248]$\\
$2$ & $1.9374;[1.0941,2.9775]$ & $.1009;[.0626,.1373]$\\
\bottomrule
\end{tabular}

\end{table}
Reducing $\alpha$ delays and reduces damage, but it also reduces the payoff on
real text. Uniform scaling therefore moves along a stability--adaptation
frontier rather than separating useful and harmful writes.

\subsection{Real read-only context helps because of its content}
Every eighth generation slot is replaced by a read-only slot, leaving 92
generated-text writes in every condition. The model receives no update in that slot. A
matched no-op condition skips the same slots, so the comparison separates the
content read in attention from the pause in generation.

\begin{table}[ht]
\caption{Read-only content at a matched write schedule. Negative differences
improve prediction relative to skipping the slot.}
\label{tab:anchor-content-expanded}
\centering\small
\begin{tabular}{lrr}
\toprule
Read-only content & Final NLL & Difference from no-op [95\% CI]\\
\midrule
No-op & 6.4042 & 0\\
Self-generated & 6.7214 & .3172 [-.4775,1.1963]\\
Random tokens & 7.3969 & .9927 [-.2699,2.0992]\\
Shuffled real & 5.5680 & -.8362 [-1.7643,.0324]\\
Fixed Generation text & 5.7912 & -.6130 [-1.5127,.2200]\\
Real text & 4.8349 & -1.5693 [-2.3011,-.8433]\\
\bottomrule
\end{tabular}

\end{table}
Real text lowers final NLL by $1.5693$ nats relative to the no-op schedule and
outperforms random, shuffled-real, and Fixed Generation anchors in direct paired
comparisons. Thus real text can interrupt the loop through attention
without being stored in weights. In the separate policy comparison, read-only
real context ends $1.3979$ nats above
Writes Off (Table~\ref{tab:attt-expanded}), so it is a partial repair rather than
a complete selection rule.

\FloatBarrier

\section{Evaluating transfer with independent evidence}
\label{app:settlement-completed}
The mechanism experiments show that source fit does not certify transfer. This
appendix turns that observation into a controlled question: if a candidate
state predicts independent real text better, can it be retained without
recreating the closed loop? We define the sequential rule, report its evidence
requirements and scale results, and then separate validation from weaker
updates and better context.
The hybrid policy evaluates a pending candidate on arriving real
text before learning from that text, and uses retained real text when no new
external text arrives. Each candidate is compared with the currently accepted
state using the same validation context. Accepted candidates update that
reference before the next candidate is evaluated. The decision does not
inspect the candidate's source label, but selecting trusted real validation
text does require knowing which evidence is real. Evaluation passages are separate from
admission text. The policy uses external validation; it does not claim
that validation itself is a new learning principle.

\subsection{Sequential validation and commitment}
\label{app:settlement-algorithm}
Algorithm~\ref{alg:settlement} describes the decision rule for an ordered
list of pending fast-weight increments. Each increment was computed from
its source chunk before the validation passage became eligible; the passage
must not be the text used to produce that increment. In the hybrid policy,
new external text is preferred and retained real text supplies a fallback.
Outcome-evaluation passages are not used to choose updates.

\begin{algorithm}[!ht]
\small
\begin{algorithmic}[1]
\Require Accepted fast weights $W$; ordered pending increments $P$;
eligible new real text $q_{\rm new}$; retained real-text bank $B$;
fixed validation context $C$
\State $q\gets q_{\rm new}$ if available; otherwise eligible text from $B$
\If{no eligible $q$ is available}
    \State \Return $(W,P)$ \Comment{Wait; buffer capacity is configuration-specific}
\EndIf
\State $S\gets W$
\For{each $\delta$ in $P$, in proposal order}
    \State $\ell_0\gets \Call{ReadOnlyNLL}{q,S,C}$
    \State $\ell_1\gets \Call{ReadOnlyNLL}{q,S+\delta,C}$
    \If{$\ell_1\leq\ell_0$}
        \State $S\gets S+\delta$ \Comment{Next candidate is judged against accepted changes}
    \EndIf
    \State Remove this candidate from $P$; record losses and decision
\EndFor
\State Commit $W\gets S$
\State \Return $(W,P)$
\end{algorithmic}
\caption{Settlement processes pending fast-weight updates sequentially.}
\label{alg:settlement}
\end{algorithm}

\textsc{ReadOnlyNLL} restores the same attention context, positions, buffers,
and other carried state for both evaluations and leaves the live stream
unchanged. It does not train on $q$. Only accepted increments change the
committed fast weights; ordinary reading of arriving input occurs outside
this procedure. The next increment is evaluated against the accumulated
accepted state, not the original $W$. Candidate increments are not jointly
summed after separate approval against an unchanged reference.

For Quarter Dose, the proposed increment is one quarter of the clipped
generated-text update before it enters $P$. The validation comparison thus
judges the same increment that would be committed. This differs from scaling
the loss before clipping, accepting a full update and scaling it afterwards,
or decaying previously accepted weights. Retention is not part of this
ablation's validation rule.

Evidence scheduling and pending-buffer capacity are implementation choices;
the reported experiments use the schedules stated with each comparison.

\subsection{Cross-scale outcomes}

\begin{table}[ht]
\caption{Completed Settlement outcomes and later admission-log extraction.
The two scales use separate matched suites and Writes Off trajectories.}
\label{tab:settlement-scale-curated}
\centering\small
\begin{tabular}{lccc}
\toprule
Model / books & Closed Loop minus Writes Off & Settlement minus Writes Off & Generated admitted\\
\midrule
125M / 12 & 3.183 [2.106,4.253] & .067 [-.013,.169] & 22/936\\
760M / 4 & 1.263 [.799,1.618] & -.021 [-.058,.014] & 18/312\\
\bottomrule
\end{tabular}

\end{table}
At 125M, Settlement improves on Closed Loop on 11 of 12 books; at 760M,
it improves all four books. The residual intervals include zero but are not
equivalence bounds and do not establish superiority to Writes Off. Real-text
acceptance is 83\% and 100\% in the respective log summaries.

\subsection{Component ablation}
\label{app:settlement-components}
This four-book, three-seed comparison separates external validation from two
weaker controls: quarter-sized updates and periodic read-only real context.
Table~\ref{tab:components-curated} reports final NLL above Writes Off. Without
validation, each control leaves a visible gap. With Settlement, every
combination finishes within $.04$ nats of Writes Off. Adding read-only real
context to Settlement lowers NLL by a further $.057$ nats
$[.024,.089]$; the incremental effect of Quarter Dose is not resolved.

\begin{table}[ht]
\caption{Completed component ablation, 125M, four books and three seeds.
Rounded interval endpoints at zero are not equivalence statements.}
\label{tab:components-curated}
\centering\small
\begin{tabular}{lclc}
\toprule
Without validation & Excess [95\% interval] & With Settlement & Excess [95\% interval]\\
\midrule
Closed Loop & 3.06 [.67,7.02] & Settlement & .04 [.00,.09]\\
Quarter Dose & .90 [.32,1.97] & + Quarter Dose & .02 [-.01,.04]\\
Read-Only Real Context & 1.04 [.19,2.59] & + Real Context & -.02 [-.03,.00]\\
Both controls & .28 [.13,.53] & + Both controls & .00 [-.01,.02]\\
\bottomrule
\end{tabular}

\end{table}

\subsection{Admission and adaptation in one all-real stream}
\label{app:settlement-pareto}
A rule can match Writes Off by rejecting every proposal. We therefore measure
admission and adaptation together on an all-real stream. The three policies
process the same six long-stream books with three seeds. \emph{Reject All}
supplies the no-write baseline, \emph{Write All} retains every real-text
proposal, and Settlement applies the same sequential rule used above. Because
the first evaluation occurs after adaptation has begun, benefit is the policy's
first-to-last NLL improvement relative to Reject All.

\begin{table}[ht]
\caption{Settlement admits most real-text candidates and preserves their
adaptation benefit in the same matched experiment. Positive benefit is better;
intervals resample books after averaging seeds.}
\label{tab:settlement-pareto}
\centering\small\setlength{\tabcolsep}{5pt}
\begin{tabular}{lrr}
\toprule
Policy & Real admitted & Benefit [95\% interval]\\
\midrule
Reject All & $0/113$ & $0$\\
Write All & $113/113$ & $.0450\;[.0235,.0609]$\\
Settlement & $\mathbf{89/113}$ & $\mathbf{.0534\;[.0299,.0721]}$\\
\bottomrule
\end{tabular}
\end{table}

Settlement admits 78.8\% of the real-text proposals. Its paired benefit over
Write All is $.0083$ $[.0020,.0161]$. The low admission rate on self-generated
streams is therefore a property of those candidate updates, not a fixed
tendency to reject all learning.

\subsection{Preserving adaptation and validating the committed state}
\label{app:commitment}
The scheduled gates below alternate admission and evaluation passages so the
same passage does not both authorize and evaluate an update. Harm and benefit
in Table~\ref{tab:commitment-appendix} come from separate streams, not a joint
experiment such as Table~\ref{tab:settlement-pareto}. Harm uses four books (five seeds for ordinary/scheduled
gates; three for Settlement). Real-text benefit uses eleven
teacher-forced books and \emph{No Updates} as its baseline.

\begin{table}[ht]
\caption{Validate the state that will be committed. Joint commit checks each
candidate at $W$ but deploys their unevaluated sum; sequential policies judge
the accumulated state $S_i$.}
\label{tab:commitment-appendix}
\centering\small\setlength{\tabcolsep}{4pt}
\begin{tabular}{lccrr}
\toprule
Policy & Candidate judged against & State committed & $H$ ($\downarrow$) & $B$ ($\uparrow$)\\
\midrule
Ordinary writing & --- & $W+\delta_t$ & $+1.56$ & $+.034$\\
Joint commit & $W$ (each $\delta$) & $W+\sum_i\delta_i$ & $\approx0$ & $-1.093$\\
Sequential & $S_{i-1}$ & $S_i$ & $+.10$ & $+.019$\\
Settlement & $S_{i-1}$ on next $q$ & $S_i$ & $\approx0$ & $+.037$\\
\bottomrule
\end{tabular}

\end{table}

\begin{figure}[!ht]
\centering
\includegraphics[width=.96\linewidth]{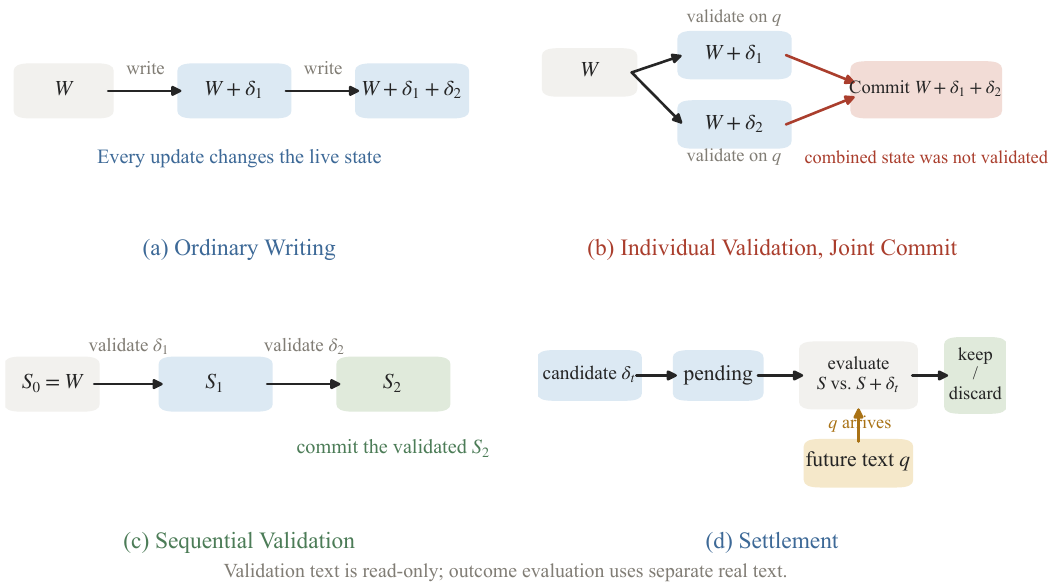}
\caption{\textbf{Four ways to apply persistent updates.}
(a) Ordinary writing immediately changes the live state. (b) Joint commit
validates candidates separately at $W$ but deploys their unevaluated sum.
(c) Sequential validation judges each candidate against the accumulated
accepted state. (d) Settlement keeps a candidate pending until arriving
external text $q$ can evaluate the state that would be committed.}
\label{fig:commitment-policies}
\end{figure}

Table~\ref{tab:commitment-appendix} reports the numerical comparison.
Positive benefit improves on No Updates; missing intervals are not inferred
from other rows.
The individual gate accepts 113/113 real candidates computed against a shared
base, then commits their combination. This committed state matches
unconditional deferred commitment and has terminal drift .0862, versus .0176
for ordinary sequential writing. Sequential validation evaluates each next
increment against the accumulated accepted state, retains about 56\% of the
ordinary real-text benefit, and avoids that large loss. This identifies a
failure of this joint application; it does not rule out alternative
batch normalization, increment recomputation, or learning-rate corrections.
Settlement accepts 98--101 of 112 real candidates. Its point
benefit is close to ordinary writing. Table~\ref{tab:settlement-pareto}
provides the matched admission--benefit comparison.

\FloatBarrier

\section{External validation when source labels are corrupted}
\label{app:online-mixed}
\label{app:provenance-noise}
Source Masking uses a real/generated label to decide whether a write should be
retained. Settlement uses the observed labels only to locate real validation
text; admission is determined by the candidate state's validation advantage
$A$. We keep the underlying 125M mixed real/generated stream fixed and
corrupt an increasing fraction of the observed labels. Each condition uses
three seeds. All Writes retains every update, and Oracle Source Masking uses
the uncorrupted labels.

\begin{table}[ht]
\caption{Corrupting source labels harms Source Masking but has little effect on
Settlement. Lower final NLL is better. The middle columns show the incorrect
decisions made by Source Masking. Bold marks comparisons whose paired 95\%
interval excludes zero in favor of Settlement.}
\label{tab:provenance-noise}
\centering\scriptsize\setlength{\tabcolsep}{3.5pt}
\begin{tabular}{rrrrr}
\toprule
Observed error & Generated kept & Real discarded & Source Masking NLL & Settlement NLL\\
\midrule
.000 & 0 & 0 & 3.8072 & 3.8196\\
.053 & 4 & 2 & 3.8116 & 3.8230\\
.124 & 9 & 5 & 3.8129 & 3.8248\\
.212 & 15 & 9 & 3.8377 & 3.8286\\
.389 & 26 & 18 & 3.8584 & \textbf{3.8215}\\
.619 & 41 & 29 & 3.9242 & \textbf{3.8270}\\
.876 & 64 & 35 & 4.0204 & \textbf{3.8243}\\
1.000 & 73 & 40 & 4.2544 & \textbf{3.8244}\\
\bottomrule
\end{tabular}
\end{table}

With accurate labels, Source Masking has the lower endpoint NLL. As errors
increase, it retains generated writes and discards real writes; its NLL rises
from 3.8072 to 4.2544. Settlement stays between 3.8196 and 3.8286 and becomes
lower in mean NLL from 21.2\% observed error. The paired Settlement-minus-
Masking intervals exclude zero at the evaluated error rates .389, .619, .876,
and 1.000; the mean difference at .212 is not resolved by its interval.

The stable result is not caused by a constant admission policy. At 100\%
observed-label error, regrouping Settlement decisions by true source gives
120/120 real candidates admitted and 0/219 self-generated candidates admitted;
regrouping the same decisions by the corrupted labels reverses these counts.
The validation advantage $A$ therefore continues to separate the candidate
states even when the labels used to locate validation text are wrong.

\section{WebShop feedback and task-metric validation}
\label{app:agent-credit}
The main experiments evaluate independent real-text prediction. WebShop \citep{webshop} asks
whether the same feedback path also changes task completion. The experiment
uses persistent rank-16 LoRA updates \citep{hu2022lora} to Qwen3-4B
\citep{qwen3} rather than TTT-E2E.
The candidate LoRA update is optimized on prompt--action pairs whose action
targets are generated by the agent. All candidates therefore receive the same
source label. A source-only rule can either reject all of them, giving Writes
Off, or retain all of them, giving Closed Loop; it cannot select among them.
Other agent TTT methods can instead learn from environment observations or
trajectory summaries, as discussed in Section~\ref{sec:scope}.
\emph{Writes Off} uses the initial model and retains no action updates.
\emph{Closed Loop} retains updates, so the adapted model chooses the actions
that form later training episodes. \emph{Fixed Generation} uses the same
update rule, but a frozen copy $W_0$ supplies every action for the learner.
\emph{Settlement} accumulates updates in a temporary model while the committed
model acts. It retains the candidate only when its mean validation reward is
strictly higher. Final evaluations use the learner with writing disabled.

\begin{table}[!ht]
\caption{WebShop outcomes across five paired seeds. Exact success is mean
$\pm$ SD; reward is the mean score. Writes Off is a shared deterministic
reference. Retained counts pool 25-episode blocks across seeds.}
\label{tab:agent-settlement-formal}
\centering\small\setlength{\tabcolsep}{8pt}
\begin{tabular}{lccc}
\toprule
Policy & Reward & Exact success & Retained blocks\\
\midrule
Writes Off & $.3309$ & $.1600$ & $0$\\
Closed Loop & $.4356$ & $.1053\pm.0597$ & $180/180$\\
Fixed Generation & $.3724$ & $.1800\pm.0089$ & $180/180$\\
Settlement & $.3823$ & $\mathbf{.1853\pm.0202}$ & $10/180$\\
\bottomrule
\end{tabular}
\end{table}

The base model is frozen, and LoRA is applied to \texttt{down\_proj} in the
final quarter of layers (about 1.8M trainable parameters). Each written
episode takes two Adam steps at $10^{-5}$ on at most 16 prompt--action pairs.
Each policy processes 900 episodes and is evaluated on the same 150 goals
from the test partition \citep{webshop}. Closed Loop, Fixed Generation, and Settlement use the
same five seeds; Writes Off is the same deterministic reference in each
comparison. The environment uses a fixed product catalogue, BM25 retrieval,
and model-scored candidate actions, including template-generated search
queries. Exact success means reward $\geq1$, not merely making a purchase.

Fixed Generation raises exact success from $.1053\pm.0597$ to
$.1800\pm.0089$ and beats Closed Loop in every paired seed. Settlement
evaluates a candidate every 25 training episodes on 10 validation episodes.
Validation and final evaluation use seeds 4321 and 1234, respectively, to
select tasks from the test partition; the runner does not enforce disjoint
goal sets. It accepts 10 of 180 candidate blocks (36 per seed) and reaches exact success
$.1853\pm.0202$, again beating Closed Loop in every paired seed. The feedback
path therefore changes task behavior as well as language-model loss. These
results compare policies within this task protocol; a common source label
cannot make the distinctions supplied by validation reward.

\FloatBarrier

\clearpage
\section{ALFWorld: keeping the agent able to complete tasks}
\label{app:alfworld}
In ALFWorld \citep{alfworld}, an agent reads room descriptions and uses text
commands to complete household tasks. We ask whether learning from its own
actions harms later task completion, and whether checking updates reduces that harm.

\paragraph{Tasks and learning.}
The Qwen3.8-27B agent \citep{qwen38_27b} attempts 140 \emph{seen} tasks followed by 134
\emph{unseen} tasks, each with a 30-action limit. Each task is scored before
its update; learned weights persist between tasks. We use LoRA
\citep{hu2022lora}, an AgentBench-style harness \citep{agentbench}, and the
official ALFWorld action parser. Scales 8 and 16 multiply the clipped weight
change. Each policy uses seeds 1234, 2345, and 3456; both scales share Writes Off.

\paragraph{Three ways to handle an update.}
\emph{Writes Off} keeps the original weights. \emph{Closed Loop} keeps every
update and uses the changed model for the next task. \emph{Settlement}
accumulates proposed updates in a temporary copy. Every 25 tasks, the current
model and this copy attempt the same 20 validation tasks. Settlement keeps the
copy only if it completes more tasks; otherwise it keeps the current model.
It also checks at task 140 (validation seed 8765). A \emph{block} is a batch
of updates accepted or rejected together: 12 per seed, or 36 per scale.

\begin{figure}[!ht]
\centering
\includegraphics[width=\linewidth]{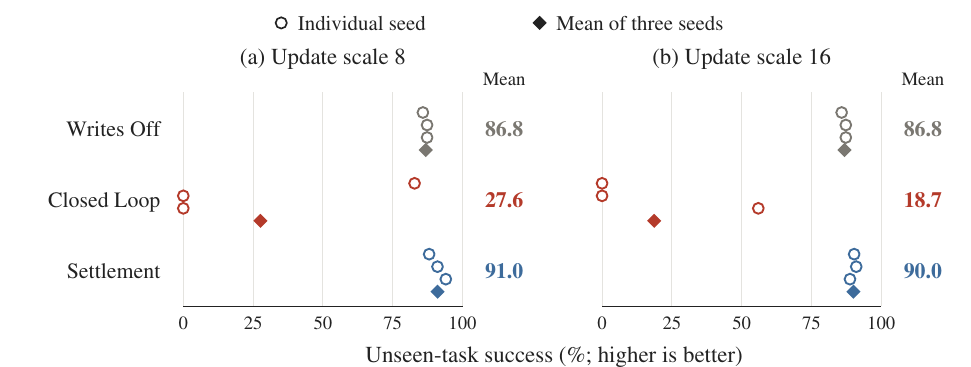}
\caption{\textbf{ALFWorld unseen-task success by update policy.} Each open
circle shows one seed's success over 134 tasks; filled diamonds and the
right-hand numbers show the three-seed mean. Points are vertically offset
to separate overlapping values. Writes Off is shared across update scales.}
\label{fig:alfworld}
\end{figure}

\begin{table}[!ht]
\caption{\textbf{ALFWorld success rates and update acceptance.} Success rates are
percentages (mean $\pm$ sample SD across three seeds). Bold marks the highest
mean per scale, including shared Writes Off, not statistical significance.
Block counts pool seeds.}
\label{tab:alfworld-settlement}
\centering\small\setlength{\tabcolsep}{3pt}
\renewcommand{\arraystretch}{1.08}
\newcommand{\alfresult}[2]{\makebox[2.0em][r]{$#1$}\,$\pm$\,\makebox[1.8em][r]{$#2$}}
\begin{tabular*}{\linewidth}{@{\extracolsep{\fill}}clrrrc@{}}
\toprule
 & & \multicolumn{3}{c}{Success rate (\%)} & Accepted\\
\cmidrule(lr){3-5}
Scale & Policy & \multicolumn{1}{c}{Overall (274)} & \multicolumn{1}{c}{Seen (140)} & \multicolumn{1}{c}{Unseen (134)} & blocks\\
\midrule
-- & Writes Off & \alfresult{82.1}{0.4} & \alfresult{\mathbf{77.6}}{0.4} & \alfresult{86.8}{0.9} & --\\
\midrule
\multirow{2}{*}{8} & Closed Loop & \alfresult{43.9}{36.3} & \alfresult{59.5}{25.6} & \alfresult{27.6}{47.8} & All updates\\
 & Settlement & \alfresult{\mathbf{83.8}}{5.0} & \alfresult{76.9}{6.9} & \alfresult{\mathbf{91.0}}{3.0} & $12/36$\\
\midrule
\multirow{2}{*}{16} & Closed Loop & \alfresult{30.7}{35.3} & \alfresult{42.1}{38.6} & \alfresult{18.7}{32.3} & All updates\\
 & Settlement & \alfresult{\mathbf{83.2}}{2.2} & \alfresult{76.7}{4.6} & \alfresult{\mathbf{90.0}}{1.1} & $7/36$\\
\bottomrule
\end{tabular*}
\end{table}

\paragraph{What changes when updates are checked?}
Halving the update scale still leaves two Closed Loop seeds with zero unseen
success. Settlement completes $88.1\%$--$94.0\%$ and retains
$19.4\%$--$33.3\%$ of update blocks. Across all 274 tasks, its mean success is
$83.2\%$--$83.8\%$, versus $82.1\%$ for Writes Off. These three-seed results
show retained updates without the observed severe failures, but do not
establish a reliable overall gain over Writes Off. Validation adds 20 tasks
per model state and decision.

\FloatBarrier

\end{document}